\documentclass{article}

\usepackage[preprint]{neurips_2026}

\usepackage[utf8]{inputenc} % allow utf-8 input
\usepackage[T1]{fontenc}    % use 8-bit T1 fonts
\usepackage{hyperref}       % hyperlinks
\usepackage{url}            % simple URL typesetting
\usepackage{booktabs}       % professional-quality tables
\usepackage{amsfonts}       % blackboard math symbols
\usepackage{nicefrac}       % compact symbols for 1/2, etc.
\usepackage{microtype}      % microtypography
\usepackage{xcolor}         % colors
\usepackage{graphicx}
\usepackage{booktabs}
\usepackage{caption}
\usepackage{subcaption}
\usepackage{colortbl}
\usepackage[most]{tcolorbox}
\usepackage{cleveref}
\usepackage{makecell}
\usepackage{multirow}
\setcitestyle{numbers,square,comma}

\newtcolorbox{qbox}[2][]{
    colback=blue!5!white,      % Background color
    colframe=blue!75!black,    % Border color
    fonttitle=\bfseries,       % Title font style
    coltitle=white,            % Title text color
    title={Question: #2},      % Title text
    enhanced,                  % Enable advanced features
    attach boxed title to top left={yshift=-2mm, xshift=2mm}, % Title positioning
    boxed title style={colback=blue!75!black},
    sharp corners=southwest,   % Mix of sharp and rounded corners
    arc=4mm,                   % Corner radius
    drop shadow,               % Add a subtle shadow
    #1                         % Allows for additional local overrides
}

\title{Spatially Localized Image Degradation Embeddings for Image Quality Assessment}

\author{%
	\begin{minipage}{0.35\textwidth}
	\centering
	\textbf{Krishna Srikar Durbha} \\
	\normalfont The University of Texas at Austin\\
	\end{minipage}
	\hfill
	\begin{minipage}{0.25\textwidth}
	\centering
	\textbf{Hassene Tmar} \\
	\normalfont Meta Platforms, Inc.\\
	\end{minipage}
	\hfill
	\begin{minipage}{0.25\textwidth}
	\centering
	\textbf{Ping-Hao Wu} \\
	\normalfont Meta Platforms, Inc.\\
	\end{minipage}
	\\
	\\
	\vspace{0.5em}
	\begin{minipage}{0.33\textwidth}
	\centering
	\textbf{Ioannis Katsavounidis} \\
	\normalfont Meta Platforms, Inc.\\
	\end{minipage}
	\hfill
	\begin{minipage}{0.33\textwidth}
	\centering
	\textbf{Alan C. Bovik} \\
	\normalfont University of Colorado Boulder \\
	\end{minipage}
}

\begin{document}

\maketitle
\begin{abstract}
	Self-supervised learning (SSL) currently drives state-of-the-art performance in no-reference image quality assessment (NR-IQA). However, standard SSL pipelines uniformly apply synthetic distortions across the entire image field, which can limit their sensitivity to spatially localized and co-occurring degradations encountered in real-world content. In this work, we empirically expose this representational ``blind spot'' across existing state-of-the-art encoders, demonstrating their reduced sensitivity to spatially bounded image degradations. To bridge this gap, we introduce \textbf{S}patial \textbf{L}ocalized \textbf{I}mage \textbf{D}egradation \textbf{E}mbeddings for Image Quality Assessment (SLIDE-IQA). SLIDE-IQA employs a dual-branch Vision Transformer framework that injects spatially bounded degradations into a contrastive pretraining objective. To handle the spatial complexity of these degradations, we introduce a Threshold-Bounded Exclusion Mechanism, a representational design choice that resolves structural conflicts arising from spatially localized distortions to ensure the latent space respects both degradation type and spatial scale. Finally, we show that SLIDE-IQA's synthetic-only pretraining significantly improves sensitivity to localized distortions, while achieving competitive performance on NR-IQA benchmarks against existing SSL NR-IQA models.
\end{abstract}

% Import Sections
\section{Introduction}
\label{sec:introduction}
No-Reference Image Quality Assessment (NR-IQA) algorithms are a core component of modern visual computing pipelines, serving as a critical metric for image compression, restoration, generation, and multimedia delivery. The primary objective of NR-IQA is to accurately predict human perceptual judgments by measuring visual deviations from the natural image manifold caused by distortions such as noise, blur, and compression artifacts, all without access to a pristine reference image. NR-IQA remains the most practically relevant, yet highly challenging, paradigm for evaluating perceptual quality ``in the wild.''

The emergence of advanced imaging pipelines such as region-of-interest (RoI) semantic compression, foveated rendering, and Generative AI is gradually shifting the nature of image distortion. In such settings, artifacts can be non-uniform and spatially localized rather than image-wide. Consequently, robust assessment of quality ``in the wild'' increasingly calls for models that are sensitive to spatially bounded distortions, in addition to globally distributed ones.

While early NR-IQA approaches relied on natural scene statistics (NSS) and statistical models \cite{BRISQUE, NIQE}, recent progress has been driven by highly parameterized convolutional and transformer-based architectures. However, the scalability of these fully supervised models is constrained by their reliance on annotated data. Collecting dense, reliable human perceptual annotations, or Mean Opinion Scores (MOS), at scale is prohibitively expensive, highly subjective, and difficult to generalize across emerging distortion types. To bypass this empirical bottleneck, the field has increasingly pivoted toward self-supervised learning (SSL) to extract robust, generalizable perceptual representations from massive corpora of unannotated images.

Current SSL pipelines for NR-IQA employ contrastive learning objectives by applying synthetic distortions uniformly across the entire image \cite{CONTRIQUE, ReIQA, ARNIQA, GRepQ, TRIQA}. While these globally pretrained models achieve high rank-correlation scores on many image-level quality benchmarks, this uniform application can diverge from the spatial structure of real-world degradations. We argue that the assumption of global spatial homogeneity does not always reflect the localized nature of many modern visual artifacts. Because existing SSL encoders exclusively observe image-wide degradations during pretraining, they are not incentivized to capture spatial variance, leaving them less sensitive to spatially localized and co-occurring degradations. Recently, authors have begun exploring this deficiency \cite{SHDIQA}, but a critical roadblock remains: standard NR-IQA benchmarks (e.g., KonIQ, FLIVE) inherently conflate global and localized errors within a single Mean Opinion Score (MOS). This makes it empirically impossible to explicitly isolate and measure an encoder's spatial sensitivity.

\begin{figure}
	\centering
	\includegraphics[width=0.9\textwidth]{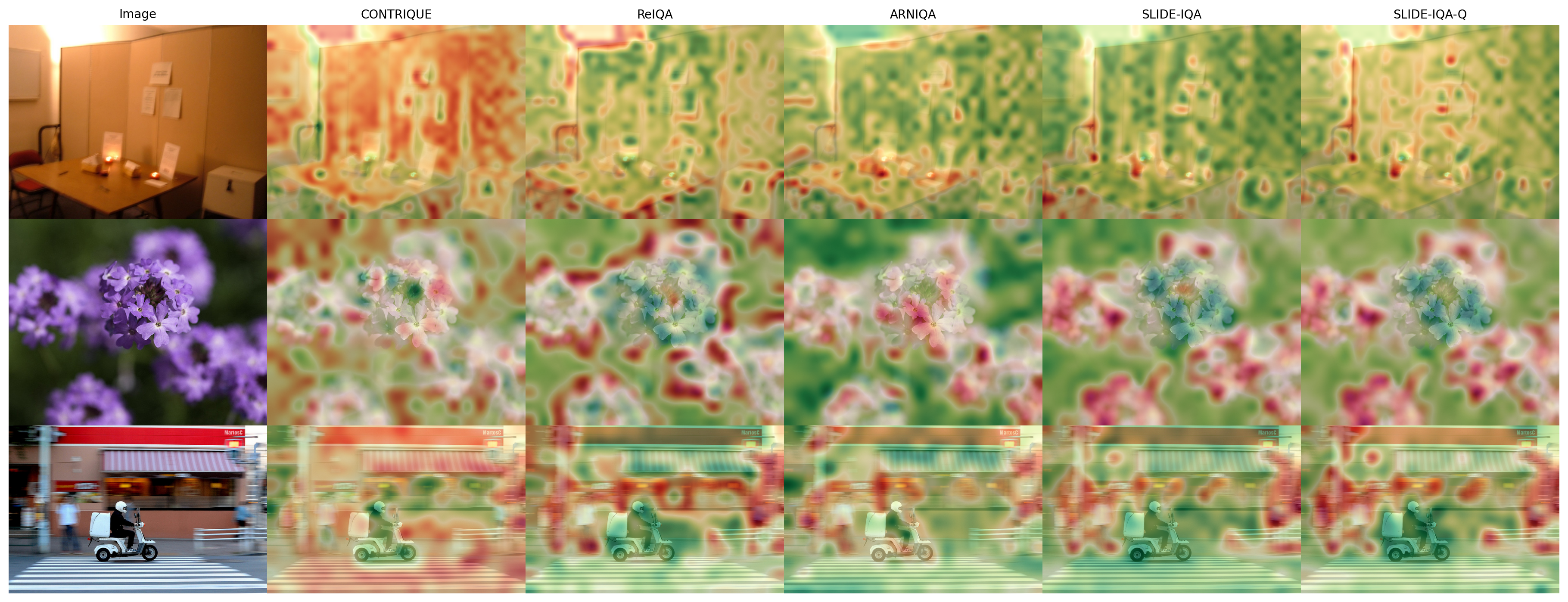}
	\caption{Patch-level quality predictions using various NR-IQA models on UGC images sourced from the KonIQ \cite{KonIQ} and FLIVE \cite{FLIVE} datasets.}
	\label{fig:patch_heatmaps}
\end{figure}

\textbf{The Blind Spot: Identifying a Representational Limitation.} To bypass the limitations of standard benchmarks, we construct a controlled diagnostic probing testbed. Rather than proposing a new dataset, we utilize this testbed as a scientific instrument to strictly isolate the spatial variable. Through linear probing, we empirically expose a representational limitation: state-of-the-art SSL encoders exhibit significant accuracy drops when detecting distortions confined to local patches or when multiple distinct degradations co-occur within a single image. This generalization gap reveals a limitation in current SSL pretraining methodologies and motivates a pretraining paradigm that incorporates localized spatial awareness.

\textbf{Towards Spatially Disentangled Representations.} To address this, we introduce Spatial Localized Image Degradation Embeddings for Image Quality Assessment (SLIDE-IQA). SLIDE-IQA utilizes a dual-branch Vision Transformer architecture to extract complementary semantic and perceptual representations. To train the quality branch, we inject localized degradations into the pretraining process by applying spatially bounded distortion masks alongside global distortions. To handle the spatial complexity of these masks, we introduce a Threshold-Bounded Exclusion Mechanism, ensuring the latent space respects both the physical degradation type and its spatial scale. The following contributions summarize our work:
\begin{itemize}
	\item We empirically expose a ``blind spot'' in current NR-IQA encoders, demonstrating via linear probing that models exclusively trained on global distortions struggle to capture localized and co-occurring degradations.
	\item We propose SLIDE-IQA, a novel dual-branch self-supervised framework that explicitly disentangles semantic and perceptual representations. By incorporating spatially bounded degradations into the pretraining procedure, we incentivize the perceptual encoder to isolate localized degradations.
	\item We adapt the contrastive learning objective to resolve structural conflicts arising from localized degradations. Specifically, we incorporate a Threshold-Bounded Exclusion Mechanism (TBEM) as a representational design choice, ensuring the learned latent space respects both degradation type and spatial scale.
	\item Through comprehensive experiments, we demonstrate that SLIDE-IQA's synthetic-only pretraining with explicit spatial bounding significantly improves the detection of complex, localized distortions, while achieving competitive performance on standard NR-IQA benchmarks against existing SSL NR-IQA models.
\end{itemize}
\section{A Diagnostic Study of Spatial Sensitivity in SSL NR-IQA Encoders}
\label{sec:diagnostic-study-spatial-sensitivity}
To empirically substantiate the representational limitations hypothesized in Section \ref{sec:introduction}, we design a linear probing framework across four state-of-the-art SSL NR-IQA image encoders: CONTRIQUE \cite{CONTRIQUE}, ReIQA \cite{ReIQA}, ARNIQA \cite{ARNIQA}, and TRIQA \cite{TRIQA}. Our objective is to determine whether feature spaces learned exclusively from globally applied synthetic distortions inherently generalize to spatially localized and compositionally complex degradations. To measure the robustness of the latent space without confounding downstream adaptation, the base encoders remain frozen, and the evaluation is modeled as a multi-label distortion classification via a single linear layer. In prior methods that utilize two encoders for content and quality representations, we conduct experiments only employing the quality encoder (e.g., ReIQA-Q refers to the quality encoder of ReIQA), as it is the component responsible for learning distortion representations.

Standard IQA datasets lack the dense, localized bounding-box annotations required to explicitly isolate local versus global degradation perception. To address this, we construct a custom diagnostic testbed (also referred to as the diagnostic test dataset for simplicity). Because evaluating all combinations of masking topologies across our full pretraining suite of distortions \cite{CONTRIQUE, ReIQA} presents a combinatorial explosion, we isolate eight highly representative distortion types commonly encountered in real-world scenarios: Gaussian Blur, Motion Blur, Saturation, Contrast, JPEG Compression, Gaussian Noise, Noise Color Map, and Pixel Jitter. To respect the physical properties of these artifacts, degradations such as JPEG Compression and sensor-level noise (Gaussian Noise, Noise Color Map, Pixel Jitter) are strictly applied globally. Conversely, optical or motion-based distortions (Gaussian Blur, Motion Blur, Contrast, Saturation) are applied both globally and locally.

We utilize 140K pristine images from the KADIS-700K dataset \cite{KADIS}, allocating 135K for training the probes and 5K for testing. To generate diverse evaluation samples, each holdout image undergoes random cropping and degradation 15 times, yielding 75K images per test set. We construct five distinct evaluation sets to systematically isolate and evaluate spatial sensitivity:
\begin{itemize}
    \item \textbf{Global Distortion}: A single synthetic distortion applied uniformly across the entire image.
    \item \textbf{Local[0.25, 0.5] + 1 Mask}: A single distortion applied to a localized spatial mask, where the mask dimensions are randomly bounded between 25\% and 50\% of the image dimensions (width or height).
    \item \textbf{Local[0.5, 0.75] + 1 Mask}: A single distortion applied to a larger spatial mask, bounded between 50\% and 75\% of the image dimensions.
    \item \textbf{Local[0.25, 0.50] + 2 Masks}: Distortions are randomly sampled and applied to two distinct spatial masks, each bounded between 25\% and 50\% of the image dimensions, simulating scenarios with multiple co-occurring localized degradations.
    \item \textbf{Global Composition}: A composition of up to two independently sampled synthetic distortions applied uniformly across the image.
\end{itemize}

\begin{table*}[t]
\centering
\caption{\textbf{Quantitative Evaluation of Spatial Entanglement via Linear Probing.} Subset Accuracy, Precision, and Recall are reported across three progressive training regimens. While models excel in their respective training distributions, they struggle significantly to generalize to localized spatial degradations. Our proposed SLIDE-IQA-Q model demonstrates improved performance compared to existing SSL encoders.}
\label{tab:quantitative_probing}
\begin{subtable}{0.49\textwidth}
    \centering
    \caption{\textbf{Global Distortion}: Linear probe trained exclusively on uniform global degradations.}
    \resizebox{\linewidth}{!}{
    \begin{tabular}{l | ccc | ccc | ccc}
        \toprule
        \multirow{2}{*}{\textbf{Encoder}} & \multicolumn{3}{c|}{\makecell{\textbf{Global}\\\textbf{Distortion}}} & \multicolumn{3}{c|}{\makecell{\textbf{Local[0.25, 0.50]}\\\textbf{1 Mask}}} & \multicolumn{3}{c}{\makecell{\textbf{Local[0.5, 0.75]}\\\textbf{1 Mask}}} \\
        \cmidrule(lr){2-4} \cmidrule(lr){5-7} \cmidrule(lr){8-10}
        & Acc & Prec & Rec & Acc & Prec & Rec & Acc & Prec & Rec \\
        \midrule
		CONTRIQUE       & 80.3 & 92.0 & 82.1 & \underline{14.7} & 52.0 & \underline{14.8} & \textbf{26.5} & 74.4 & \textbf{26.6} \\
		ReIQA-Q         & \underline{88.0} & \underline{97.6} & \underline{88.3} & 9.7 & \underline{60.6} & 9.7 & 24.2 & \underline{85.3} & 24.2 \\
		ARNIQA          & 77.7 & 90.9 & 78.8 & 14.3 & 47.6 & 14.3 & 22.4 & 59.7 & 22.5 \\
		TRIQA-Q         & 75.2 & 92.8 & 76.9 & 8.3 & 44.5 & 8.7 & 17.4 & 65.5 & 17.9 \\
		SLIDE-IQA-Q & \textbf{91.9} & \textbf{98.9} & \textbf{92.0} & \textbf{18.1} & \textbf{88.1} & \textbf{18.1} & \underline{26.0} & \textbf{91.9} & \underline{26.0} \\
        \bottomrule
    \end{tabular}
}
\end{subtable}
\hfill
\begin{subtable}{0.49\textwidth}
    \centering
    \caption{\textbf{Global Composition}: Linear probe trained on compositions up to two global distortions.}
    \resizebox{\linewidth}{!}{
    \begin{tabular}{l | ccc | ccc | ccc}
        \toprule
        \multirow{2}{*}{\textbf{Encoder}} & \multicolumn{3}{c|}{\makecell{\textbf{Global}\\\textbf{Composition}}} & \multicolumn{3}{c|}{\makecell{\textbf{Local[0.25, 0.50]}\\\textbf{1 Mask}}} & \multicolumn{3}{c}{\makecell{\textbf{Local[0.5, 0.75]}\\\textbf{1 Mask}}} \\
        \cmidrule(lr){2-4} \cmidrule(lr){5-7} \cmidrule(lr){8-10}
        & Acc & Prec & Rec & Acc & Prec & Rec & Acc & Prec & Rec \\
        \midrule
		CONTRIQUE       & 25.5 & 90.7 & 58.7 & \underline{15.8} & \underline{53.1} & \underline{16.7} & \underline{26.4} & 67.9 & \underline{27.5} \\
		ReIQA-Q         & \underline{47.0} & \underline{95.9} & \underline{72.2} & 8.4 & 52.3 & 8.8 & 23.1 & \underline{79.9} & 24.3 \\
		ARNIQA          & 30.4 & 90.2 & 61.6 & 13.2 & 45.1 & 13.8 & 20.9 & 57.3 & 22.0 \\
		TRIQA-Q         & 34.5 & 94.0 & 63.4 & 6.9 & 43.8 & 7.3 & 16.5 & 64.7 & 17.7 \\
		SLIDE-IQA-Q & \textbf{55.6} & \textbf{96.0} & \textbf{77.4} & \textbf{41.0} & \textbf{91.2} & \textbf{41.5} & \textbf{57.5} & \textbf{86.9} & \textbf{79.3} \\
        \bottomrule
    \end{tabular}
}
\end{subtable}

\vspace{3mm}

\begin{subtable}{\textwidth}
\centering
\caption{\textbf{Global + Local[0.25, 0.75] with 1 Mask}: Linear probe trained on a mixture of uniform global and single-region localized degradations.}
\resizebox{\textwidth}{!}{
    \begin{tabular}{l | ccc | ccc | ccc | ccc}
    \toprule
        \multirow{2}{*}{\textbf{Encoder}} & \multicolumn{3}{c|}{\textbf{Global Distortion}} & \multicolumn{3}{c|}{\textbf{Local [0.25, 0.50] + 1 Mask}} & \multicolumn{3}{c|}{\textbf{Local [0.50, 0.75] + 1 Mask}} & \multicolumn{3}{c}{\textbf{Local[0.25, 0.50] + 2 Masks}} \\
        \cmidrule(lr){2-4} \cmidrule(lr){5-7} \cmidrule(lr){8-10} \cmidrule(lr){11-13}
        & Accuracy & Prec & Recall & Accuracy & Prec & Recall & Accuracy & Prec & Recall & Accuracy & Prec & Recall \\
        \midrule
		CONTRIQUE       & 75.7 & 92.0 & 77.3 & 30.3 & 84.2 & 30.9 & 56.7 & 91.2 & 59.2 & 14.3 & 92.5 & 27.4 \\
		ReIQA-Q         & \underline{85.1} & \underline{96.6} & \underline{85.5} & \underline{43.8} & \underline{86.9} & \underline{44.6} & \underline{70.3} & \underline{94.8} & \underline{71.0} & \underline{19.5} & \underline{94.4} & \underline{37.1} \\
		ARNIQA          & 73.0 & 91.4 & 74.0 & 20.6 & 69.6 & 20.9 & 39.0 & 84.4 & 41.1 & 9.6 & 84.1 & 20.7 \\
		TRIQA-Q         & 68.3 & 90.3 & 70.2 & 24.1 & 72.3 & 25.1 & 47.2 & 84.9 & 50.3 & 11.8 & 87.5 & 22.6 \\
		SLIDE-IQA-Q & \textbf{90.3} & \textbf{98.9} & \textbf{90.4} & \textbf{87.7} & \textbf{99.1} & \textbf{87.7} & \textbf{89.6} & \textbf{99.3} & \textbf{89.6} & \textbf{25.9} & \textbf{99.9} & \textbf{54.8} \\
    \bottomrule
    \end{tabular}
}
\end{subtable}
\end{table*}

We define three training regimens for the linear probes to systematically evaluate spatial composition: (1) \textbf{Global Distortion}, trained exclusively on uniform global distortions; (2) \textbf{Global Composition}, trained on combinations of up to two global distortions; and (3) \textbf{Global + Local[0.25, 0.75] with 1 Mask}, trained on a mixture of global distortions and single-region localized distortions. These training paradigms allow us to systematically evaluate the generalization capabilities and serve as a basis to determine whether the latent space can effectively capture localized degradations without explicit exposure during training. Appendix provides samples from the diagnostic testbed.

Tables \ref{tab:quantitative_probing}(a-c) present the results from our linear probing experiments across the three training regimens. When trained exclusively on global distortions (i.e., mirroring CONTRIQUE and ReIQA-Q), all models demonstrate strong performance on global degradations but experience significant drops on localized distortions. For smaller masks (25-50\% of image dimensions), accuracy drops as low as 8.3\% for TRIQA-Q and 9.7\% for ReIQA-Q --- a 70-80 absolute point decrease relative to global performance. Performance marginally improves with larger masks but remains substantially below global baselines. Similarly, models trained on compositions of global distortions (ARNIQA, TRIQA) show the same vulnerability, suggesting that exposure to complex global combinations does not transfer to localized degradations.

When trained on a mixture of global and localized distortions, models exhibit slight degradation in global performance but measurable improvements on localized distortions, with ReIQA-Q showing the strongest gains. However, localized performance remains notably below global baselines. When evaluated on two distortions applied to separate spatial masks, all models experience significant accuracy drops. These results substantiate the representational blind spot hypothesized in Section \ref{sec:introduction} and indicate a need for pretraining that explicitly incorporates spatial-awareness of degradations. In the next section, we introduce SLIDE-IQA, a novel dual-branch self-supervised framework designed to address this critical limitation.
\section{Method}
\label{sec:method}
\begin{figure*}[t]
    \centering
    \includegraphics[width=\textwidth]{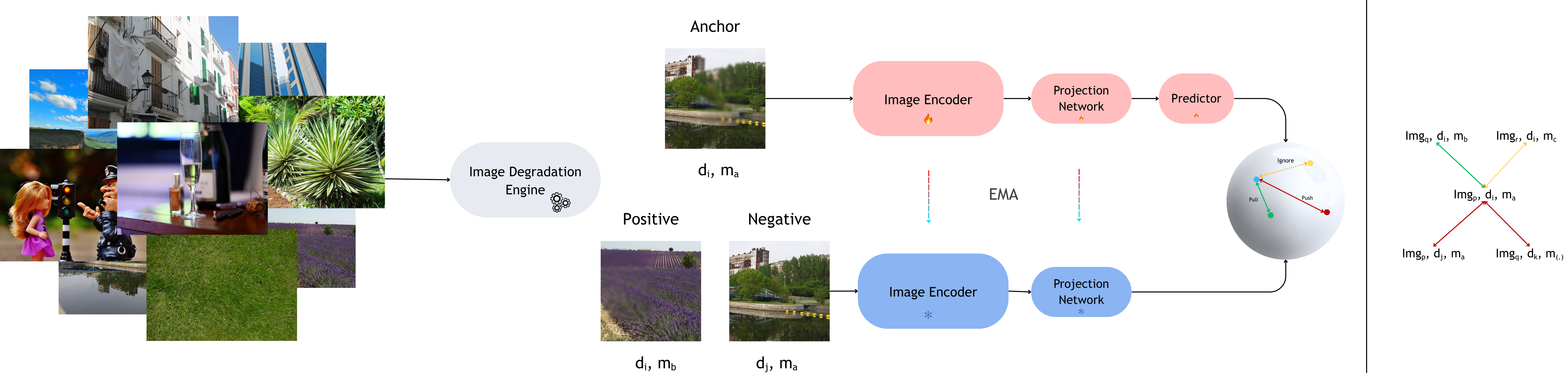}
    \caption{Overview of the proposed framework to pretrain the perceptual encoder of SLIDE-IQA.}
    \label{fig:method}
\end{figure*}

To overcome the blind spot of existing NR-IQA image encoders, we introduce Spatial Localized Image Degradation Embeddings for Image Quality Assessment (SLIDE-IQA). SLIDE-IQA utilizes two separate image encoders to extract distinct semantic and perceptual representations. Our primary contribution lies in the pretraining of the perceptual encoder, which is optimized using a novel self-supervised objective that explicitly incorporates spatially localized degradations alongside standard global distortions. Figure \ref{fig:method} shows the overall architecture of our proposed pretraining framework. \footnote{The code and the diagnostic testbed will be released upon publication.}

\subsection{Degradation Engine}
To train the perceptual encoder to recognize spatially localized artifacts, we introduce a spatially-aware degradation engine. Unlike standard pipelines that apply uniform distortions across the entire image, our engine generates spatially bounded regional masks and restricts specific degradations to those bounded areas, accurately simulating localized anomalies. We draw upon the comprehensive suite of image distortions established in prior NR-IQA literature \cite{CONTRIQUE, ReIQA, TRIQA}, encompassing 26 distinct synthetic image distortions evaluated across 5 severity levels. To maintain physical plausibility, we partition these degradations into two categories:
\begin{itemize}
    \item \textbf{Strictly Global Distortions:} Artifacts resulting from holistic image processing, transmission bottlenecks, or sensor limitations (e.g., JPEG compression, Gaussian noise, bilinear resizing) are applied uniformly across the entire image.
    \item \textbf{Localizable Distortions:} Optical aberrations, focal anomalies, and illumination shifts (e.g., Gaussian blur, motion blur, saturation changes) are applied either globally or strictly within the stochastically sampled spatial bounding boxes.
\end{itemize}
During pretraining, the engine dynamically samples from these categories to generate single global distortions, isolated regional defects, global distortion compositions, or complex compositions with both global and localized distortions. More details on the specific distortions and their partitioning are provided in the Appendix.

\subsection{Pretraining Objective}
To learn degradation-aware representations without relying on image semantics, we employ an asymmetric student-teacher architecture based on MoCo v3 \cite{MoCov3}. Our goal is to explicitly disentangle perceptual artifacts from image content, avoiding the semantic shortcuts common in standard contrastive learning, where models simply learn to recognize objects.

At each training iteration, we sample two random pristine images, $I_A$ and $I_B$. Our degradation engine then samples two degradation operations, $\mathcal{D}_1$ and $\mathcal{D}_2$ (comprising the distortion type and severity), alongside two spatial masks, $\mathcal{M}_1$ and $\mathcal{M}_2$. We construct a training triplet as follows:
\begin{align}
	x_a &= \mathcal{A}(I_A, \mathcal{D}_1, \mathcal{M}_1) \\
	x_p &= \mathcal{A}(I_B, \mathcal{D}_1, \mathcal{M}_2) \\
	x_n &= \mathcal{A}(I_A, \mathcal{D}_2, \mathcal{M}_1)
\end{align}
where $\mathcal{A}$ is the function that applies the specified degradation within the designated spatial mask. The anchor ($x_a$) and negative ($x_n$) share identical underlying image content and spatial bounds, but differ in degradation. Conversely, the anchor and positive ($x_p$) share an identical physical degradation, but differ entirely in their underlying semantic content and spatial location. Unlike standard instance-discrimination contrastive learning, where all other samples in a batch are treated as negatives, this triplet construction explicitly groups pairs based on physical degradations. However, because the number of discrete degradation types is small compared to the dataset size, multiple samples in a large batch frequently share the same degradation. Treating these identical degradations as negative pairs creates false-negative collisions. To systematically resolve this, we map each sample to a discrete degradation class label $d$ and a normalized spatial mask ratio vector $m \in [0, 1]^{\text{K}}$ where $K$ is the number of spatial masks applied in the degradation and each component $m_k$ represents the fraction of image pixels covered by the $k$-th mask.

If two images share the same degradation class but possess vastly different spatial mask ratios, a structural conflict occurs. Treating them as negative pairs incorrectly penalizes the model for recognizing the same physical distortion, while treating them as positive pairs forces the model to ignore spatial scale entirely. To resolve this, we introduce a geometric threshold $\gamma$. For an anchor sample $i$ in a batch $\mathcal{B}$ with degradation class $d_i$ and mask vector $m_i$, we partition the remaining samples $j \in \mathcal{B}$ into three distinct sets using the $L_1$ norm ($\|\cdot\|_1$):
\begin{enumerate}
    \item \textbf{Positives $\mathcal{P}(i)$:} Samples sharing the exact degradation class with a spatial footprint bounded by the threshold: $\{j \in \mathcal{B} \setminus \{i\} \mid d_i = d_j \text{ and } \|m_i - m_j\|_1 \le \gamma\}$.
    \item \textbf{Negatives $\mathcal{N}(i)$:} Samples with a fundamentally different degradation class: $\{j \in \mathcal{B} \mid d_i \neq d_j\}$.
    \item \textbf{Ignored $\mathcal{V}(i)$:} Samples sharing the exact degradation class but with a spatial variance exceeding the threshold: $\{j \in \mathcal{B} \mid d_i = d_j \text{ and } \|m_i - m_j\|_1 > \gamma\}$.
\end{enumerate}
By enforcing these relationships, the model is explicitly incentivized to isolate perceptual quality from image semantics. It learns to group identical distortions together (via positives) while distinguishing between distinct distortion families (via negatives), all while remaining sensitive to the spatial scale of the degradation (via the ignored set). For a given anchor $i$, yielding an $L_2$-normalized query vector $\mathbf{q}_i$ from the student and key vectors $\mathbf{k}$ from the teacher, the objective is defined as:
\begin{align}
 \mathcal{L}_i = \frac{-1}{|\mathcal{P}(i)|} \sum_{p \in \mathcal{P}(i)} \log \frac{\exp(\mathbf{q}_i \cdot \mathbf{k}_p / \tau)}{\sum_{j \in \mathcal{P}(i) \cup \mathcal{N}(i)} \exp(\mathbf{q}_i \cdot \mathbf{k}_j / \tau)}
\end{align}
where $\tau$ is the temperature hyperparameter, $\mathbf{q}_i$ is the projection from student encoder and $\mathbf{k}_j$ is the key vector from the teacher encoder. Critically, the ignored set $\mathcal{V}(i)$ is completely excluded from both numerator and denominator. By construction, the denominator sums only over $\mathcal{P}(i) \cup \mathcal{N}(i)$, ignoring samples with structural conflicts, i.e., the same degradation class but vastly different spatial scales. This mechanism allows the network to learn robust spatial representations without conflating different scales of the same degradation.

\subsection{No-Reference Quality Prediction}
While the perceptual branch is explicitly trained to isolate degradation features, authentic NR-IQA evaluation requires understanding both the severity of a distortion and its interaction with underlying image content. We employ DINOv3 to extract semantic features and the perceptual encoder pretrained via our proposed framework to extract features explicitly sensitive to localized degradations. Following pretraining, the weights of both encoders are frozen. We then evaluate on standard NR-IQA benchmarks using a linear regressor trained on concatenated features from both branches to predict the Mean Opinion Score (MOS). This ensures that the performance gains can be attributed to the quality of the learned representations rather than the regression model.
\section{Experimental Results}
\label{sec:experiments}
\subsection{Experimental Setup and Implementation Details}
\textbf{Pretraining Data:} We utilize 135,000 pristine images randomly sampled from the KADIS-700k dataset \cite{KADIS} to pretrain the perceptual encoder. A disjoint set of 5,000 pristine images is reserved exclusively for validation and as source images for our diagnostic spatial probe (Section \ref{sec:diagnostic-study-spatial-sensitivity}). This strict separation ensures the model is never exposed to the diagnostic source images during pretraining. Unlike prior methods \cite{CONTRIQUE, ReIQA} that pretrain directly on authentic user-generated content (UGC), SLIDE-IQA is exposed exclusively to synthetic degradations during pretraining, forcing the model to learn fundamental degradation patterns rather than dataset-specific biases.

\textbf{Training Protocol:} The perceptual branch, initialized with DINOv3 ViT-S/16, is pretrained for 100 epochs. We employ a global batch size of 384, optimized using AdamW \cite{AdamW} with a base learning rate of $1 \times 10^{-4}$ scaled based on the batch size and a cosine decay schedule following a 1000-step linear warmup. The contrastive temperature is set to $\tau = 0.1$. The geometric threshold for resolving structural conflicts is set to $\gamma = 0.05$. The teacher encoder is updated using an exponential moving average. More details on the training hyperparameters, data augmentations, and computational resources are provided in the Appendix.

\textbf{Degradation Engine Settings:} During pretraining, images are subjected to synthetic distortions with probability $p = 0.95$. To simulate complex artifacts, the engine dynamically constructs random compositions consisting of one to two distortions. For localized degradations, a single rectangular mask is generated per distortion, with its size uniformly sampled between $25\%$ and $75\%$ of the image dimensions. Furthermore, to ensure a robust representation of holistic artifacts, global distortions remain in the sampling pool, and any generated localized mask is stochastically dropped with probability $p = 0.20$, forcing the selected distortion to be applied globally across the entire image.

\textbf{NR-IQA Benchmarks:} We evaluate the downstream generalization of SLIDE-IQA on a comprehensive suite of eight standard NR-IQA benchmarks. This encompasses four authentic distortion datasets characterized by complex, co-occurring in-the-wild degradations (KonIQ-10k \cite{KonIQ}, CLIVE \cite{CLIVE}, FLIVE \cite{FLIVE}, and SPAQ \cite{SPAQ}), and four synthetic distortion datasets (LIVE-IQA \cite{LIVE-IQA}, CSIQ-IQA \cite{CSIQ-IQA}, TID-2013 \cite{TID-2013}, and KADID-10k \cite{KADID}). We follow the standard evaluation protocols for each dataset, using the provided training and testing splits where applicable, following \cite{CONTRIQUE, ReIQA}. Except for FLIVE, we report the median PLCC and SRCC values across 10 random seeds. Appendix provides additional details on the evaluation protocols for each dataset. We compare performance against the state-of-the-art SSL methods, including CONTRIQUE \cite{CONTRIQUE}, ReIQA \cite{ReIQA}, ARNIQA \cite{ARNIQA}, and TRIQA \cite{TRIQA}, along with prior supervised methods \cite{DBCNN,PQR,FLIVE,HyperIQA,TReS,MUSIQ} and handcrafted baselines \cite{BRISQUE,CORNIA}.

\subsection{Performance on NR-IQA Benchmarks}
\begin{table*}[t]
	\centering
	\scriptsize
	\setlength{\tabcolsep}{4pt}
	\renewcommand{\arraystretch}{1.02}
	\resizebox{\textwidth}{!}{%
	\begin{tabular}{@{}l *{16}{c}@{}}
		\midrule
		\multirow{3}{*}{Method} & \multicolumn{8}{c}{Authentic Distortions --- ``Images in the Wild''} & \multicolumn{8}{c}{Synthetic Distortions} \\
		\cmidrule(lr){2-9} \cmidrule(lr){10-17}
		& \multicolumn{2}{c}{KonIQ} & \multicolumn{2}{c}{CLIVE} & \multicolumn{2}{c}{FLIVE} & \multicolumn{2}{c}{SPAQ} & \multicolumn{2}{c}{LIVE-IQA} & \multicolumn{2}{c}{CSIQ-IQA} & \multicolumn{2}{c}{TID-2013} & \multicolumn{2}{c}{KADID} \\
		\cmidrule(lr){2-3} \cmidrule(lr){4-5} \cmidrule(lr){6-7} \cmidrule(lr){8-9} \cmidrule(lr){10-11} \cmidrule(lr){12-13} \cmidrule(lr){14-15} \cmidrule(lr){16-17}
		& SRCC & PLCC & SRCC & PLCC & SRCC & PLCC & SRCC & PLCC & SRCC & PLCC & SRCC & PLCC & SRCC & PLCC & SRCC & PLCC \\
		\midrule
		BRISQUE         & 0.665 & 0.681 & 0.608 & 0.629 & 0.288 & 0.373 & 0.809 & 0.817 & 0.939 & 0.935 & 0.746 & 0.829 & 0.604 & 0.694 & 0.528 & 0.567 \\
		CORNIA          & 0.780 & 0.795 & 0.629 & 0.671 & -     & -     & 0.709 & 0.725 & 0.947 & 0.950 & 0.678 & 0.776 & 0.678 & 0.768 & 0.516 & 0.558 \\
		DB-CNN          & 0.875 & 0.884 & 0.851 & 0.869 & 0.554 & 0.652 & 0.911 & 0.915 & 0.968 & 0.971 & 0.946 & 0.959 & 0.816 & 0.865 & 0.851 & 0.856 \\
		PQR             & 0.880 & 0.884 & \underline{0.857} & \underline{0.882} & -     & -     & -     & -     & 0.965 & 0.971 & 0.872 & 0.901 & 0.740 & 0.798 & -     & -     \\
		PaQ-2-PiQ       & 0.870 & 0.880 & 0.840 & 0.850 & 0.571 & 0.623 & -     & -     & -     & -     & -     & -     & -     & -     & -     & -     \\
		HyperIQA        & 0.906 & 0.917 & \textbf{0.859} & \textbf{0.882} & 0.535 & 0.623 & 0.916 & 0.919 & 0.962 & 0.966 & 0.923 & 0.942 & 0.840 & 0.858 & 0.852 & 0.845 \\
		TReS			& 0.915 & 0.928 & 0.846 & 0.877 & 0.554 & 0.625 & -     & -     & 0.969 & 0.968 & 0.922 & 0.942     & 0.863 & 0.883 & 0.859 & 0.858 \\
		MUSIQ           & \textbf{0.916} & \underline{0.928} & -     & -     & \textbf{0.646} & \textbf{0.739} & \underline{0.917} & \underline{0.921} & -     & -     & -     & -     & -     & -     & -     & -     \\
		\midrule
		CONTRIQUE       & 0.894 & 0.906 & 0.845 & 0.857 & 0.580 & 0.641 & 0.914 & 0.919 & 0.960 & 0.961 & 0.942 & 0.955 & 0.843 & 0.857 & \textbf{0.934} & \textbf{0.937} \\
		ReIQA-C/MoCov2	& 0.896 & 0.912 & 0.808 & 0.844 & 0.588 & 0.699 & 0.902 & 0.908 & 0.867 & 0.858 & 0.766 & 0.824 & 0.658 & 0.736 & 0.601 & 0.656 \\
		ReIQA-Q 		& 0.861 & 0.885 & 0.806 & 0.824 & 0.584 & 0.590 & 0.900 & 0.910 & \textbf{0.971} & \textbf{0.972} & 0.944 & \underline{0.964} & 0.844 & 0.880 & 0.885 & 0.892 \\
		ReIQA           & 0.914 & 0.923 & 0.840 & 0.854 & \underline{0.645} & \underline{0.733} & \textbf{0.918} & \textbf{0.925} & \underline{0.970} & \underline{0.971} & \underline{0.947} & 0.960 & 0.804 & 0.861 & 0.872 & 0.885 \\
		ARNIQA          & -     & -     & -     & -     & 0.595 & 0.671 & 0.905 & 0.910 & 0.966 & 0.970 & \textbf{0.962} & \textbf{0.973} & 0.880 & \underline{0.901} & 0.908 & 0.912 \\
		TRIQA           & 0.915 & 0.926	& 0.837 & 0.871 & 0.567 & 0.653	& 0.911 & 0.914 & -     & - 	& -		& -		& -		& - 	& -     & - \\
		\midrule
		SLIDE-IQA-C/DINOv3 & 0.887 & 0.909 & 0.784 & 0.807 & 0.571 & 0.685 & 0.892 & 0.896 & 0.925 & 0.928 & 0.653 & 0.764 & 0.666 & 0.757 & 0.514 & 0.534 \\
		SLIDE-IQA-Q  & 0.854 & 0.869 & 0.690 & 0.710 & 0.516 & 0.576 & 0.885 & 0.890 & 0.955 & 0.961 & 0.940 & 0.954 & \underline{0.881} & 0.892 & \underline{0.932} & \underline{0.934} \\
		SLIDE-IQA    & \underline{0.915} & \textbf{0.930} & 0.825 & 0.851 & 0.610 & 0.708 & 0.913 & 0.918 & 0.963 & 0.968 & 0.925 & 0.947 & \textbf{0.899} & \textbf{0.911} & 0.924 & 0.927 \\
		\bottomrule
		\end{tabular}
	}
	\caption{Performance of various NR-IQA methods on authentic and synthetic distortion datasets. The top section lists supervised methods, while the middle and bottom sections present SSL methods and our proposed SLIDE-IQA, respectively. The best performance for each dataset is highlighted in bold, while the second-best is underlined.}
	\label{tab:values-from-papers}
\end{table*}

Table \ref{tab:values-from-papers} summarizes the performance of SLIDE-IQA against prior state-of-the-art methods across eight standard NR-IQA benchmarks, with baseline metrics from prior works \cite{CONTRIQUE, ReIQA, ARNIQA, TRIQA}. It can be observed that fully supervised methods that rely on annotations achieve strong performance on authentic datasets but show high variability on synthetic data. In contrast, SSL methods utilizing global synthetic distortions during pretraining excel on synthetic datasets but struggle to match the performance of supervised models on authentic degradations. SLIDE-IQA demonstrates superior performance on KonIQ \cite{KonIQ} and TID-2013 \cite{TID-2013} while remaining competitive on other benchmarks. While methods such as CONTRIQUE \cite{CONTRIQUE} and ReIQA \cite{ReIQA} explicitly pretrain on authentic User-Generated Content (UGC), SLIDE-IQA remains competitive on authentic datasets despite being strictly trained on synthetic data during pretraining, demonstrating that our spatially bounded synthetic pretraining provides a good learning signal for authentic distortions.

Furthermore, it can also be observed that the two branches of SLIDE-IQA exhibit complementary performance patterns. The quality encoder (SLIDE-IQA-Q) achieves stronger performance on synthetic datasets than the semantic branch (SLIDE-IQA-C, using DINOv3). Conversely, the semantic branch exhibits stronger performance on authentic distortion datasets such as FLIVE \cite{FLIVE} and CLIVE \cite{CLIVE}. This mirrors the behavior observed in prior dual-branch architectures \cite{ReIQA}, demonstrating the complementary nature of the two branches and the necessity of leveraging both semantic and perceptual representations for NR-IQA evaluation across diverse distortion patterns. Figure \ref{fig:patch_heatmaps} shows patch-level quality predictions from various NR-IQA models on UGC images sourced from the KonIQ \cite{KonIQ} and FLIVE \cite{FLIVE} datasets. It can be observed that SLIDE-IQA-Q demonstrates good sensitivity to localized degradations without being explicitly trained on UGC images.

\subsection{Encoder Representation Alignment with Distortion Sensitivity}
While standard NR-IQA benchmarks provide primary evaluation, they are limited in spatial diversity. Specifically, synthetic distortion datasets constrain degradations globally to the entire image, failing to reflect the compositional complexity of localized artifacts. Therefore, we evaluate the spatial sensitivity of encoder representations on the diagnostic testbed introduced in Section \ref{sec:diagnostic-study-spatial-sensitivity}. Figure \ref{fig:friqa_quality_scores} plots the distribution of quality scores from full-reference IQA (FR-IQA) methods \cite{PSNR, SSIM, E-SSIM, VIF, LPIPS} across our diagnostic testbed.

\begin{minipage}{0.62\textwidth}
	While FR-IQA metrics effectively differentiate between pristine and globally distorted images, their quality scores for localized degradations remain heavily skewed toward higher quality values, indicating limited sensitivity to local spatial bounds. Among the evaluated FR-IQA methods, VIF \cite{VIF} demonstrates the highest sensitivity to localized degradations, followed by LPIPS \cite{LPIPS} and Enhanced SSIM \cite{E-SSIM}. We provide an extended analysis of FR-IQA performance in the Appendix. We employ VIF as a reference signal to evaluate the spatial sensitivity of encoder representations, since our synthetic testbed lacks human perception annotations.\\

	Table \ref{tab:friqa_quality_scores} shows cosine similarity between encoder representations of reference and distorted images against VIF scores \cite{VIF}. SLIDE-IQA-Q achieves the highest correlation across diagnostic test sets, with notable margins on localized degradations. It may also be observed from Table \ref{tab:quantitative_probing} that SLIDE-IQA-Q demonstrates higher performance in detecting the presence of localized degradations under linear probing. These findings validate our hypothesis: representations learned with explicit spatial bounding (SLIDE-IQA-Q) maintain sensitivity to localized artifacts, while globally pretrained baselines show reduced spatial sensitivity. We report an extended analysis on the image encoder and the impact of $\gamma$ in the Appendix. We also further demonstrate that the gain in performance on localized degradations does not come from architectural differences but through the pretraining paradigm.
\end{minipage}
\hfill
\begin{minipage}{0.36\textwidth}
    \centering
    \captionsetup{type=figure}
    \subcaptionbox{Global Distortions}{
        \includegraphics[width=\linewidth]{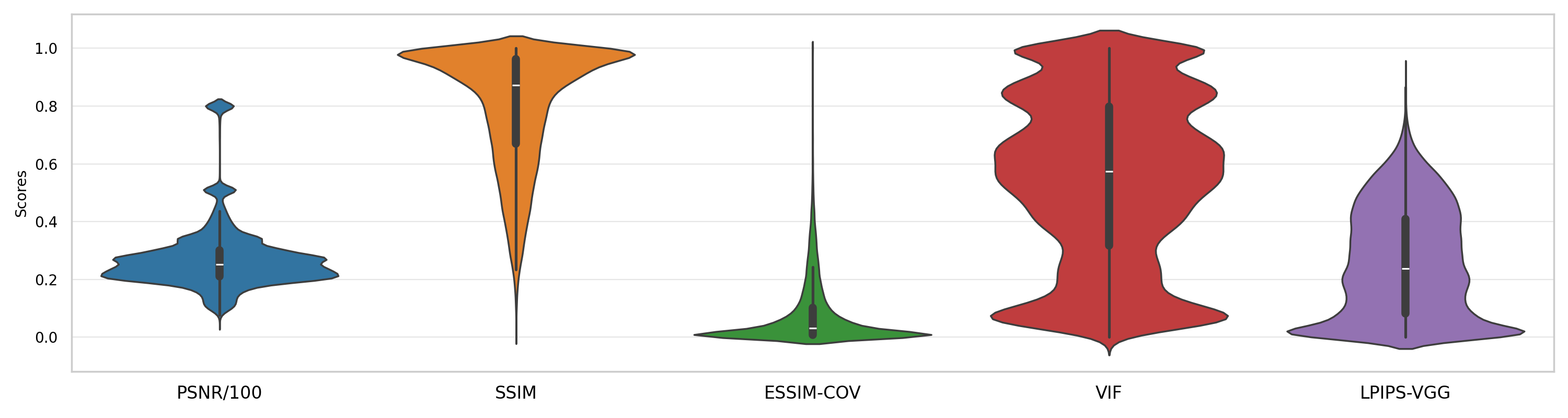}
    }\\[6pt]
    \subcaptionbox{Local[0.25, 0.50] + 1 Mask}{
        \includegraphics[width=\linewidth]{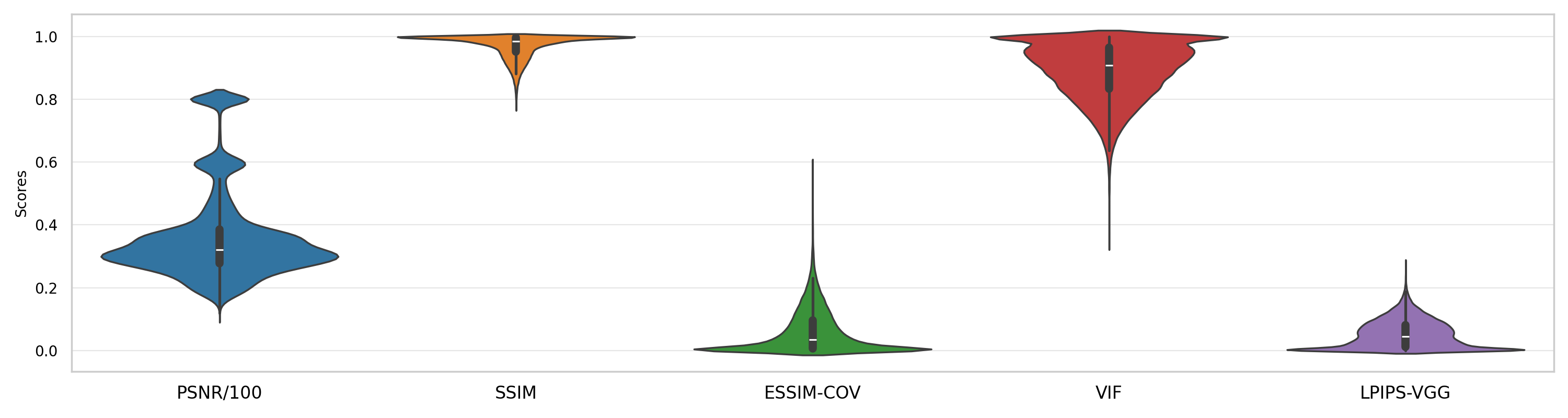}
    }\\[6pt]
    \subcaptionbox{Local[0.5, 0.75] + 1 Mask}{
        \includegraphics[width=\linewidth]{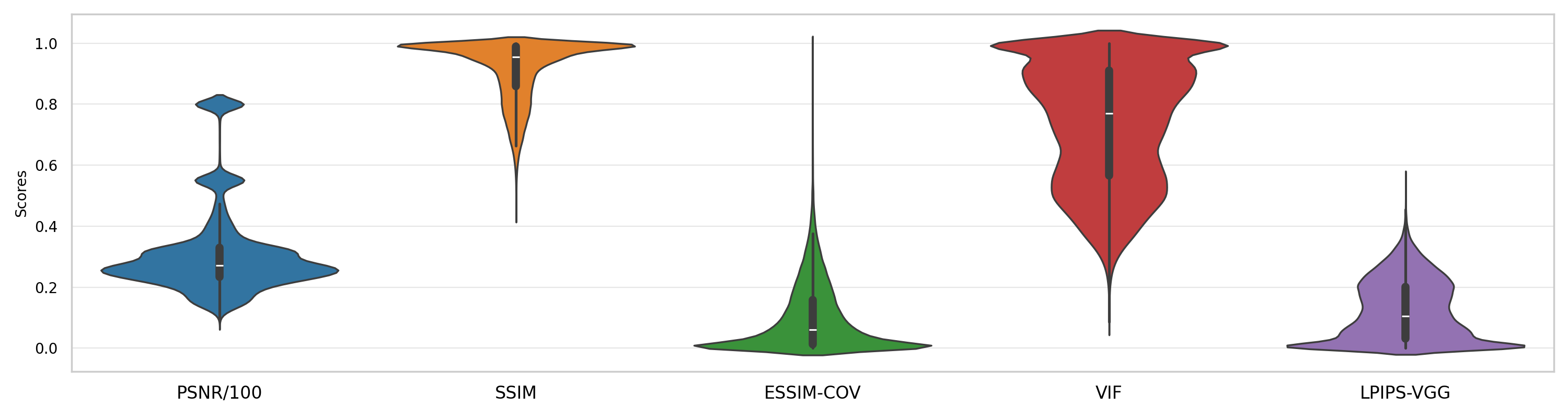}
    }\\[6pt]
    \subcaptionbox{Global Composition}{
        \includegraphics[width=\linewidth]{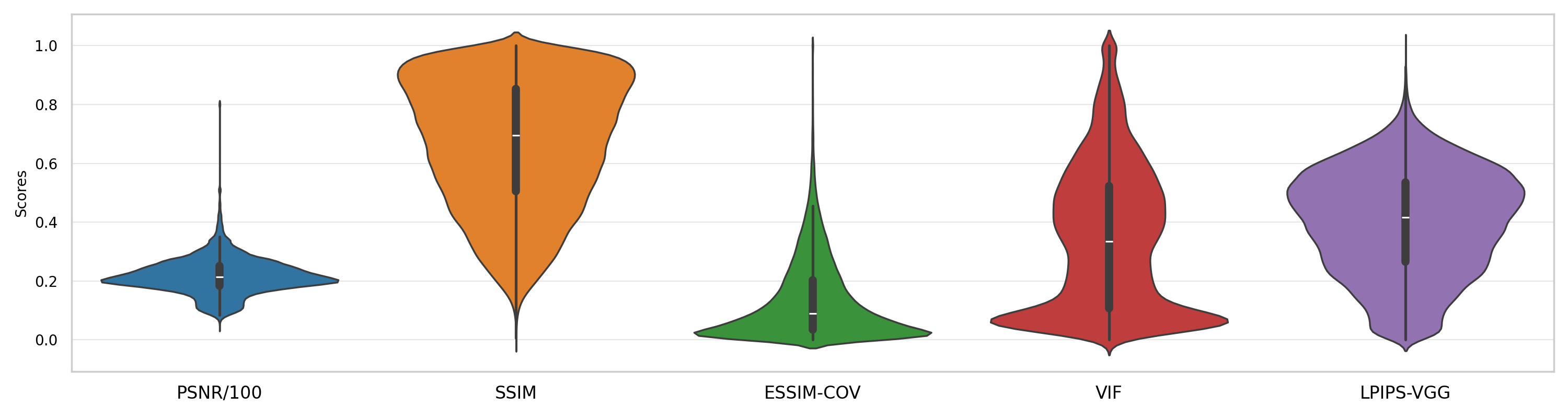}
    }
    \caption{Distribution of quality scores from various FR-IQA methods on our diagnostic test dataset.}
    \label{fig:friqa_quality_scores}
\end{minipage}

\begin{table}
    \centering
    \caption{Performance of cosine similarity between encoder representations against VIF scores on our diagnostic testbed. The best performance for each distortion type is highlighted in bold, while the second-best is underlined.}
    \label{tab:friqa_quality_scores}
    \resizebox{\textwidth}{!}{
    \begin{tabular}{l | cc | cc | cc | cc}
    \toprule
        \multirow{2}{*}{\textbf{Encoder}} & \multicolumn{2}{c|}{\textbf{Global Distortion}} & \multicolumn{2}{c|}{\textbf{Local [0.25, 0.50] + 1 Mask}} & \multicolumn{2}{c|}{\textbf{Local [0.50, 0.75] + 1 Mask}} & \multicolumn{2}{c}{\textbf{Local[0.25, 0.50] + 2 Masks}} \\
        \cmidrule(lr){2-3} \cmidrule(lr){4-5} \cmidrule(lr){6-7} \cmidrule(lr){8-9}
        & SRCC & PLCC & SRCC & PLCC & SRCC & PLCC & SRCC & PLCC \\
        \midrule
        CONTRIQUE       & 75.4 & \underline{74.6} & \underline{65.6} & \underline{47.8} & \underline{71.2} & \underline{63.4} & 53.0 & \underline{47.8} \\
        ReIQA-Q         & 66.6 & 65.1 & 59.6 & 38.3 & 65.0 & 50.3 & 48.4 & 39.0 \\
        ARNIQA          & 81.9 & \textbf{79.6} & 58.4 & 24.0 & 58.2 & 32.7 & 41.9 & 24.4 \\
        TRIQA-Q         & \underline{82.4} & 70.9 & \textbf{68.0} & 32.2 & \textbf{72.1} & 40.2 & \underline{55.9} & 33.5 \\
        SLIDE-IQA-Q & \textbf{83.2} & \textbf{79.6} & \textbf{68.0} & \textbf{60.0} & 70.0 & \textbf{64.1} & \textbf{56.5} & \textbf{53.2} \\
    \bottomrule
    \end{tabular}
    }
\end{table}

\subsection{Qualitative Analysis}
\begin{figure}
    \centering
    \begin{subfigure}{0.19\linewidth}
        \includegraphics[width=\linewidth]{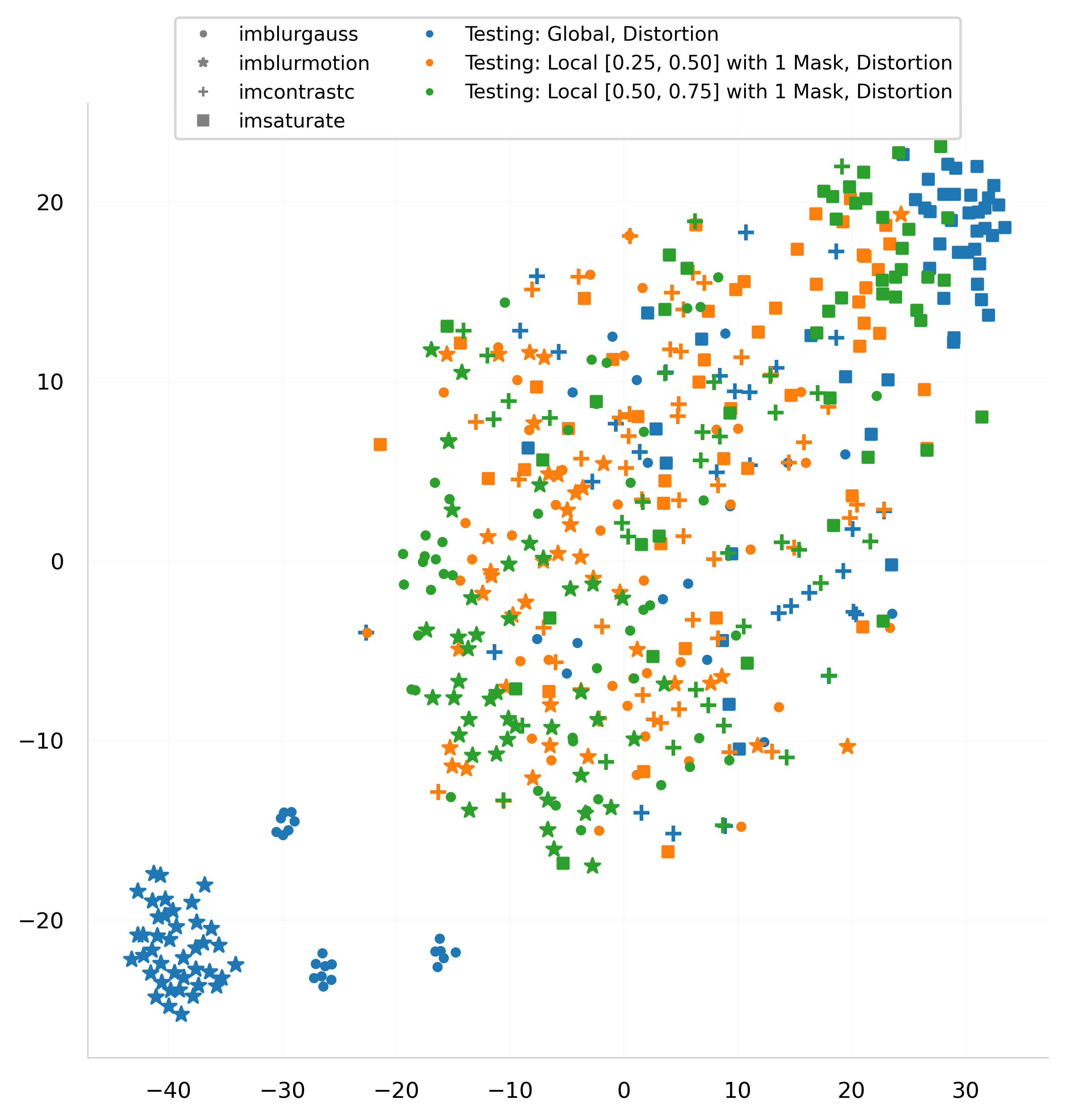}
        \caption{CONTRIQUE}
    \end{subfigure}
    \begin{subfigure}{0.19\linewidth}
        \includegraphics[width=\linewidth]{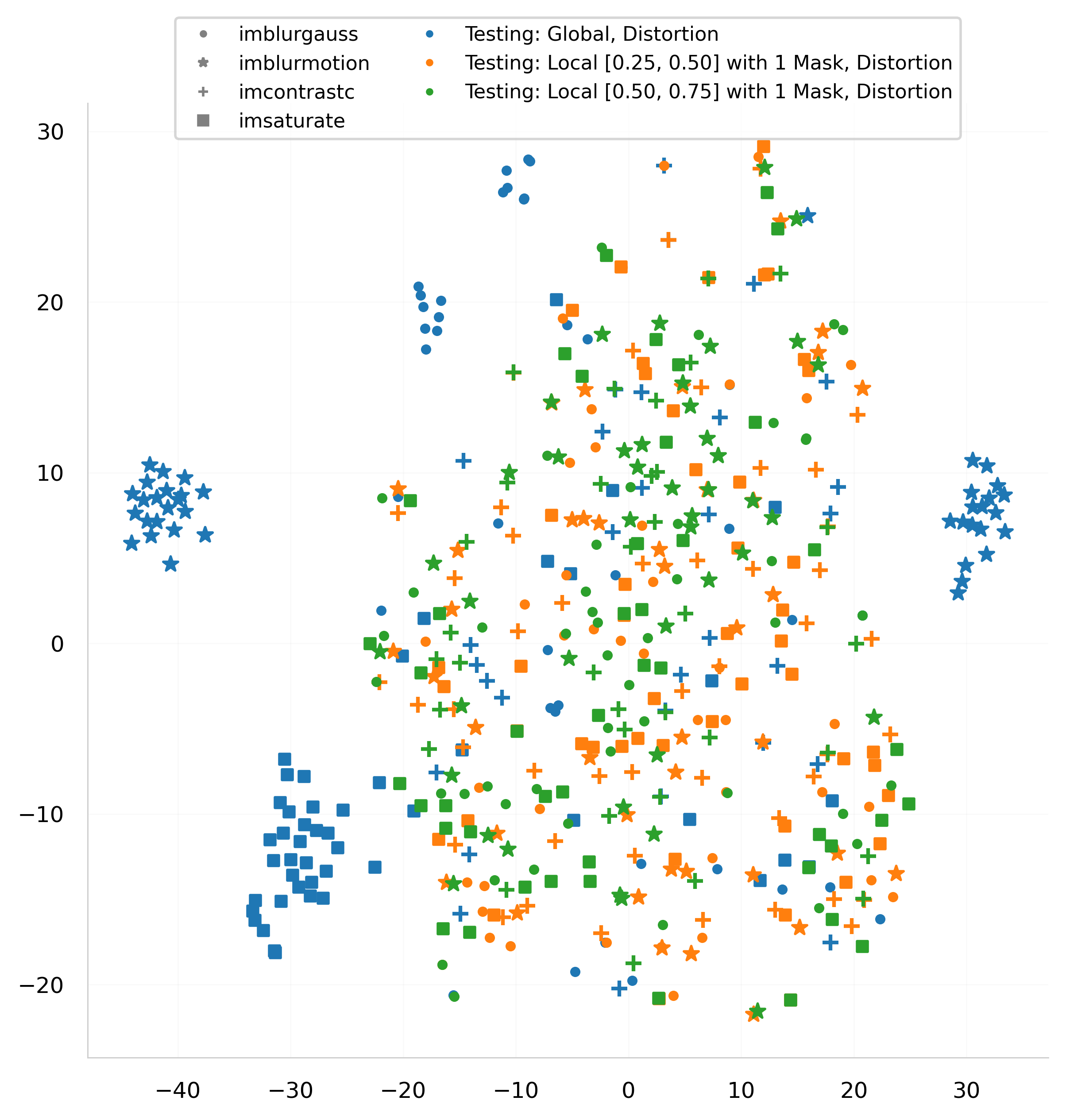}
        \caption{ReIQA-Q}
    \end{subfigure}
    \begin{subfigure}{0.19\linewidth}
        \includegraphics[width=\linewidth]{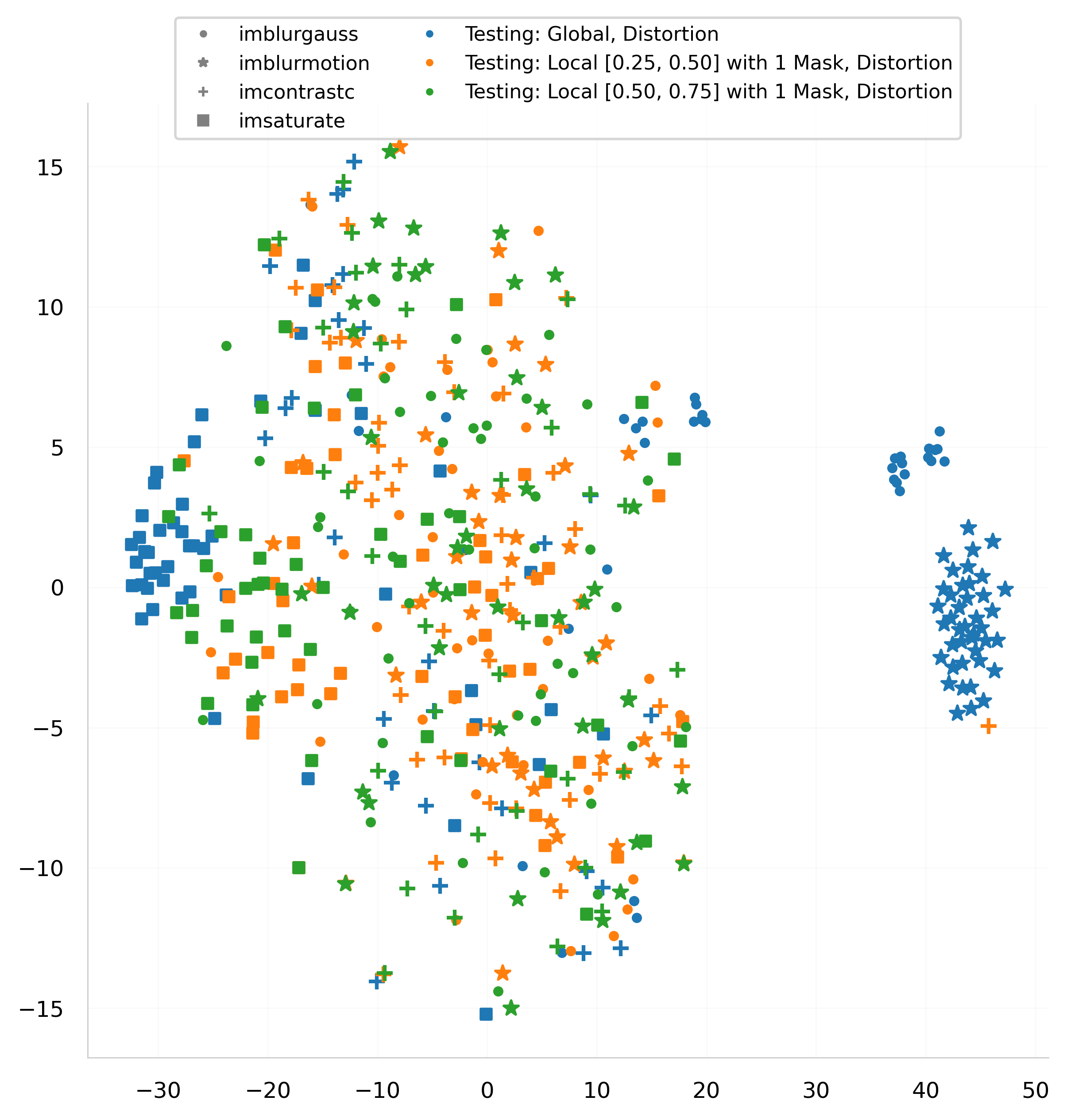}
        \caption{ARNIQA}
    \end{subfigure}
    \begin{subfigure}{0.19\linewidth}
        \includegraphics[width=\linewidth]{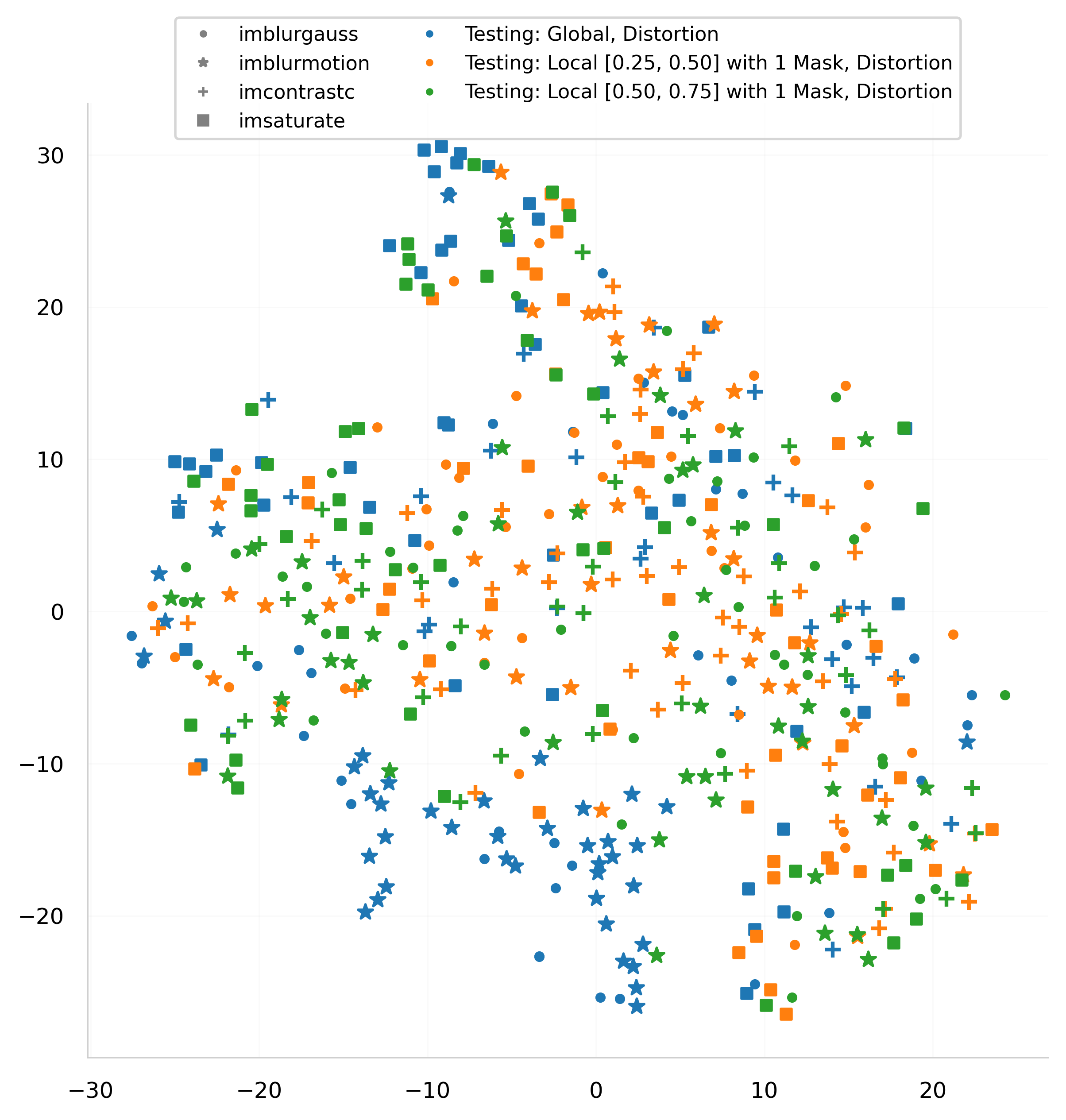}
        \caption{TRIQA-Q}
    \end{subfigure}
    \begin{subfigure}{0.19\linewidth}
        \includegraphics[width=\linewidth]{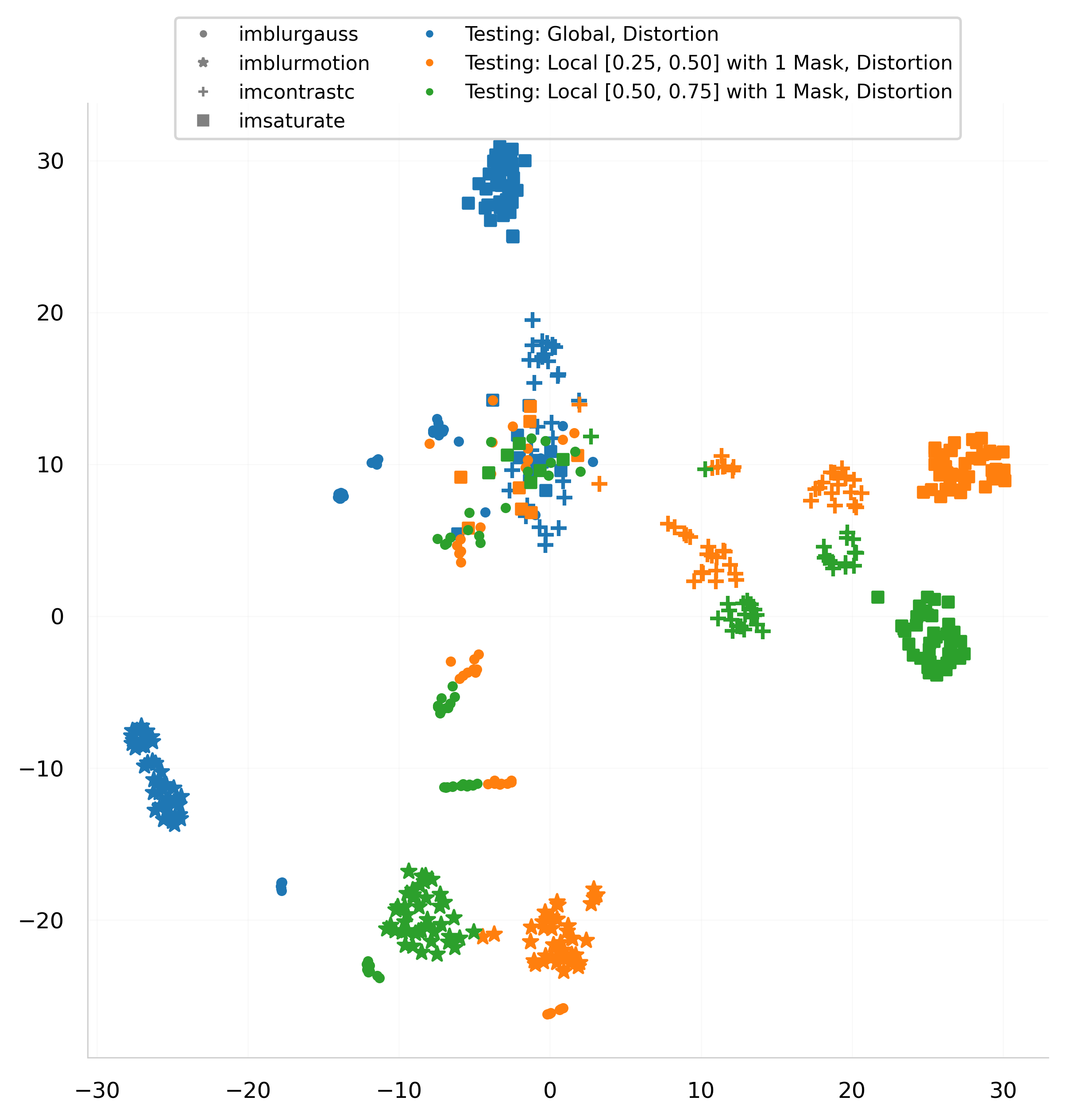}
        \caption{SLIDE-IQA-Q}
    \end{subfigure}
    \caption{t-SNE visualizations of the representations learned by various methods on samples from our diagnostic test datasets.}
    \label{fig:t_sne}
\end{figure}

Figure \ref{fig:t_sne} presents t-SNE visualizations of the latent representations extracted by various models on sample images from our diagnostic testbed. While prior methods \cite{CONTRIQUE, ReIQA, ARNIQA, TRIQA} demonstrate clear separation of the latent space for global distortions, they fail to maintain distinct clustering for localized degradations. Consequently, representations of locally distorted images show reduced separation, with samples scattering across the manifold. This visual evidence corroborates the accuracy drops empirically demonstrated in Section \ref{sec:diagnostic-study-spatial-sensitivity}. In contrast, SLIDE-IQA-Q preserves degradation-specific clustering even when artifacts are strictly spatially bounded. This qualitative analysis further supports our hypothesis about the blind spot in existing SSL encoders and highlights the effectiveness of our proposed pretraining framework in learning more robust representations for NR-IQA.

\section{Related Works}
\label{sec:related_works}
No-Reference Image Quality Assessment (NR-IQA) has evolved from handcrafted statistical models \cite{BRISQUE,NIQE,BLIINDS,DIIVINE,CORNIA,HOSA} to deep representation learning models. Supervised NR-IQA methods \cite{RankIQA,DBCNN,PQR,FLIVE,HyperIQA,TReS,MUSIQ} have traditionally focused on learning mappings from image features to human perceptual scores. However, the scarcity of annotated data has motivated the exploration of self-supervised learning (SSL) techniques that can leverage large-scale unannotated datasets. These methods \cite{CONTRIQUE, ReIQA, ARNIQA, TRIQA, QPTv2, GRepQ} leverage contrastive learning objectives to learn robust perceptual representations, ensuring the image encoder \cite{ResNet50, ViT, ConvNext} is sensitive to a diverse set of image degradations. However, these methods have been limited in their ability to capture spatially localized distortions as they primarily rely on global image transformations during pretraining. Recently, SHDIQA \cite{SHDIQA} has made strides in addressing this issue by introducing spatially localized distortions during training. However, they rely on distilling knowledge from a supervised teacher model pre-trained on annotated IQA data, meaning the framework is not fully self-supervised and remains indirectly bottlenecked by the need for expensive human-annotated Mean Opinion Scores (MOS). In contrast, our proposed SLIDE-IQA is designed to be fully self-supervised, eliminating the dependency on annotated data and allowing for more scalable and efficient pretraining. Appendix provides a more detailed discussion of related works in NR-IQA.
\section{Conclusion}
\label{sec:conclusion}
In this work, we empirically exposed the representational blind spot in existing state-of-the-art SSL NR-IQA models. By isolating the spatial variable through a rigorously controlled diagnostic probing testbed, we demonstrated that globally pretrained models exhibit significant performance drops when detecting localized distortions or handling multiple co-occurring degradations. To address this limitation, we introduced SLIDE-IQA, a dual-branch self-supervised framework designed to explicitly disentangle semantic and perceptual representations. By injecting spatially bounded degradations into the pretraining objective and resolving spatial conflicts via a Threshold-Bounded Exclusion Mechanism, SLIDE-IQA incentivizes the perceptual encoder to isolate localized artifacts. Our experiments demonstrate that our framework achieves competitive performance across standard NR-IQA benchmarks while being significantly more effective at assessing localized distortions.

\section{Limitations and Future Work}
Our diagnostic probing testbed focuses on a limited set of synthetic degradations, drawn from the standard distortions used to pretrain various SSL NR-IQA encoders. While these degradations are commonly used in the field and provide a controlled environment for isolating the spatial variable, they may not fully capture the diversity and complexity of real-world UGC distortions. Additionally, our testbed does not account for the interplay between textures and image degradations, which could be an important factor in NR-IQA performance. Future work could explore a wider range of degradation types and patterns, as well as investigate the effects of degradation on different image textures. 

\section*{Acknowledgements}
The authors thank the Texas Advanced Computing Center (TACC) at The University of Texas at Austin for providing HPC resources that have contributed to the research results reported in this paper. URL: http://www.tacc.utexas.edu.

% Appendix
% \clearpage
\appendix
\section{Appendix}
\label{sec:appendix}

\subsection{Comprehensive Review of Related Works in NR-IQA}
\label{sec:related_works_appendix}
\textbf{Early NR-IQA Methods:} No-Reference Image Quality Assessment (NR-IQA) has evolved from handcrafted statistical models to deep representation learning models. Early approaches, including BRISQUE \cite{BRISQUE} and NIQE \cite{NIQE}, extracted features from band-pass statistics of natural images to capture deviations from natural scene statistics (NSS). These models provided fundamental insights into the statistical properties of natural images and how distortions affect these properties. Later methods, such as DIIVINE \cite{DIIVINE} and BLIINDS \cite{BLIINDS}, utilized steerable pyramids and DCT coefficients, respectively, to measure traces of image degradations. CORNIA \cite{CORNIA} and HOSA \cite{HOSA} employ codebooks constructed from local patches to extract features for quality assessment. These handcrafted methods, while pioneering, struggle to generalize across the diverse distortions present in real-world User-Generated Content (UGC) due to their limited capacity to capture complex interactions between different types of distortions and their reliance on specific statistical assumptions.

\textbf{Supervised Deep Learning Methods:} With the advent of deep learning, supervised models with convolutional neural networks (CNNs) backbones have achieved significant performance gains in NR-IQA. Due to the limited availability of annotated data, these approaches often relied on pretrained weights from large-scale image classification tasks for initialization. RankIQA \cite{RankIQA} employs siamese networks to learn from pairwise rankings of image quality and is later fine-tuned on annotated datasets. DBCNN \cite{DBCNN} utilizes a dual-branch architecture with each branch pretrained on synthetic distortions and authentic distortions, respectively. PQR \cite{PQR} formulates NR-IQA by modeling the distribution of subjective opinion scores instead of just the mean opinion scores (MOS) during training, achieving faster convergence and better quality estimates. TReS \cite{TReS} leverages the self-attention mechanism to learn a non-local image representation from multi-scale features, L2 loss, and relative ranking loss. HyperIQA \cite{HyperIQA} employs a hypernetwork to predict the weights of the quality prediction model using semantic features while employing global features and local distortion features to predict quality. PaQ-2-PiQ \cite{FLIVE} employs both image and patch quality scores to demonstrate significant performance improvements in model performance. MUSIQ \cite{MUSIQ} utilizes a vision transformer (ViT) backbone to extract multi-scale features for no-reference image quality assessment. These supervised approaches, while achieving state-of-the-art performance, are bottlenecked by their reliance on expensive human annotations (Mean Opinion Scores). Recently, SHDIQA \cite{SHDIQA} addressed this issue by introducing spatially localized image degradations during training. To evaluate these regional artifacts, the framework utilizes dedicated distortion tokens and knowledge distillation to model the complex relationships between diverse local distortions. However, their reliance on a supervised teacher model pre-trained on annotated IQA data means the framework is not fully self-supervised, remaining indirectly bottlenecked by the high cost of human-annotated Mean Opinion Scores (MOS).

\textbf{Self-Supervised Learning Methods:} To address the data scarcity issue in NR-IQA, self-supervised learning (SSL) methods have emerged, leveraging contrastive objectives to learn robust perceptual representations from unannotated data. CONTRIQUE \cite{CONTRIQUE} utilizes a contrastive learning framework by employing images with synthetic and authentic distortions during the pretraining stage, posing distortion type and degree as auxiliary tasks to learn representations. ReIQA \cite{ReIQA} employs a mixture of experts framework combining two image encoders learning content and quality representations, respectively. QPTv2 \cite{QPTv2} utilizes a masked image modeling (MIM) framework and hierarchical vision transformer architecture to learn multi-scale distortion and aesthetic features. ARNIQA \cite{ARNIQA} utilizes a stronger image degradation engine employing a composition of image distortions during pretraining, enabling the model to learn more robust representations. Similar to ReIQA \cite{ReIQA}, TRIQA \cite{TRIQA} also employs a mixture of experts framework and employs image compositions and contrastive triplet-based learning during pretraining utilizing ConvNext backbone \cite{ConvNext}. These methods pretrain image encoders on synthetic image distortions applied across the entire image. However, as demonstrated in the main paper, many existing SSL methods show limitations in capturing spatially localized distortions. Our proposed SLIDE-IQA addresses this gap by introducing spatially localized distortion simulation during pretraining, enabling the model to learn representations that are sensitive to localized degradations.

\subsection{Additional Details on Insensitivity to Localized Degradations}
\begin{figure}
	\centering
	\begin{subfigure}{0.32\linewidth}
		\includegraphics[width=\linewidth]{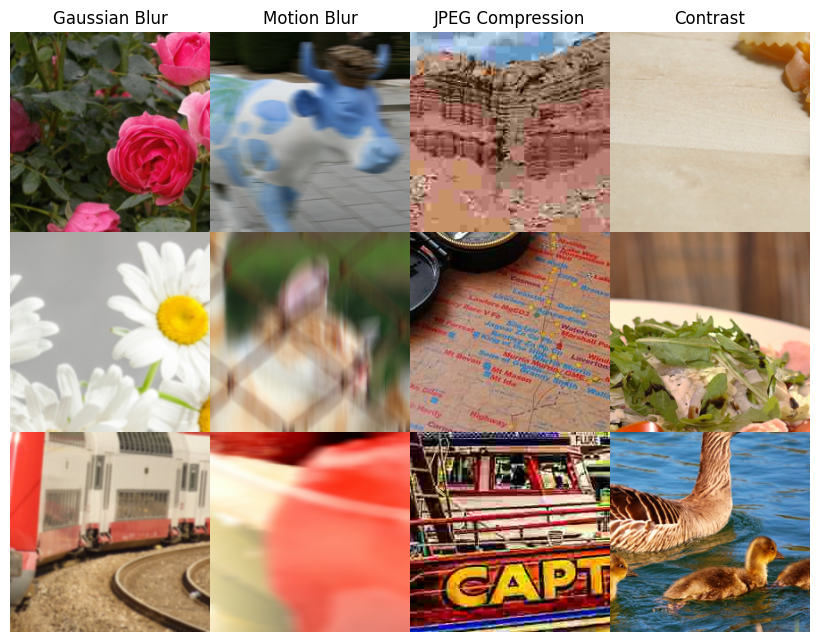}
		\caption{Global Distortion}
	\end{subfigure}
	\begin{subfigure}{0.32\linewidth}
		\includegraphics[width=\linewidth]{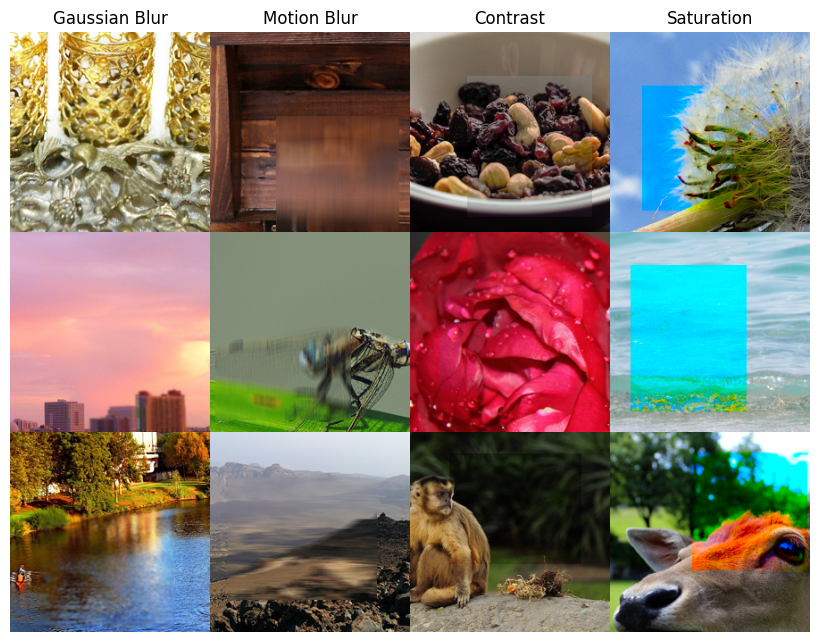}
		\caption{Local[0.5, 0.75] + 1 Mask}
	\end{subfigure}
	\begin{subfigure}{0.32\linewidth}
		\includegraphics[width=\linewidth]{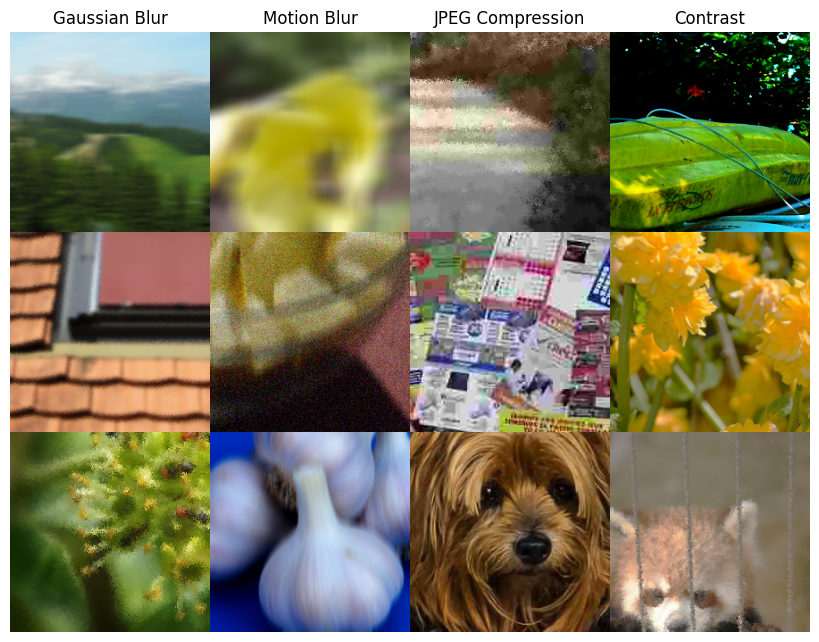}
		\caption{Global Composition}
	\end{subfigure}
	\caption{Sample images from our diagnostic probing testbed, showcasing the diversity of spatially localized and compositionally complex degradations.}
	\label{fig:sample_images}
\end{figure}

\subsubsection{Additional Details on the Degradation Engine}
In this subsection, we provide additional details regarding the degradations employed in the training of the linear probes under different training regimes.
\begin{itemize}
	\item \textbf{Global Distortion:} The images are subjected to a single global distortion 95\% of the time.
	\item \textbf{Global Composition:} The images are subjected to a single global degradation 95\% of the time. A degradation can be a single image degradation or a composition with two image distortions, with each having a 50\% chance.
	\item \textbf{Global + Local[0.25, 0.75]}: The images are subjected to a degradation 95\% of the time. The degradation can be a single global distortion or a localized distortion with a rectangular mask of size uniformly sampled between 25\% and 75\% of the image size. Additionally, we ignore the mask and apply the localized distortion globally with a 20\% probability to ensure a robust representation of holistic artifacts.
\end{itemize}

\subsubsection{Optimization and Hyperparameters}
The linear probes are trained for 50 epochs with a global batch size of 512 using Multi-Label Binary Cross Entropy (BCE) loss. We employ the AdamW \cite{AdamW} optimizer with a peak learning rate of $0.001$ and an initial weight decay of $0.01$. We employ a cosine annealing learning rate schedule incorporating a 1000-step linear warmup (starting at $1 \times 10^{-6}$ and ultimately decaying to a final minimum learning rate of $1 \times 10^{-6}$). Concurrently, we utilize a cosine weight decay schedule that gradually anneals the weight decay coefficient from $0.01$ to $0.1$ over the duration of training. To ensure training stability and prevent gradient explosions, gradient clipping is applied with a maximum norm of $10.0$. We employ the square-root-256 rule to scale the learning rate based on the batch size and the number of GPUs used for training. We employ horizontal flip and random cropping as our only data augmentations before passing the images through the degradation engine. We employ NVIDIA A100 and RTX 4090 GPUs for our experiments. The training time for each linear probe setting is around 2 to 3 hours. We employ AMD Ryzen Threadripper PRO 5975WX CPUs to generate testbed datasets, which takes around 4 to 5 hrs in total. 

\subsection{Image Distortion Bank}
We consider a comprehensive suite of 26 distinct synthetic image distortions, evaluated across 5 severity levels as established in prior NR-IQA literature \cite{CONTRIQUE, ReIQA, TRIQA}.  We partition these distortions into two categories based on their physical plausibility as localized artifacts:

\begin{minipage}{0.45\textwidth}
	\quad \textbf{Strictly Global Distortions}:
	\begin{itemize}
		\item JPEG Compression
		\item Gaussian Noise
		\item Color Map Noise
		\item Impulse Noise
		\item Multiplicative Noise
		\item Resize Bicubic
		\item Resize Linear
		\item Resize Nearest
		\item Resize Lanczos
		\item Color Shift
		\item Pixelate
		\item Non-Eccentricity
		\item Image Jitter
	\end{itemize}
\end{minipage}
\hfill
\begin{minipage}{0.45\textwidth}
	\quad \textbf{Localized Distortions}:
	\begin{itemize}
		\item Gaussian Blur
		\item Lens Blur
		\item Motion Blur
		\item Color Diffuse
		\item Color Saturation
		\item Saturate
		\item Denoise
		\item Brighten
		\item Darken
		\item Mean Shift
		\item Sharpen
		\item Contrast
		\item Color Block
	\end{itemize}
\end{minipage}

While certain strictly global distortions could conceivably manifest locally in real-world scenarios, this partitioning serves as a representative baseline for our pretraining objective. Crucially, the proposed SLIDE-IQA framework is completely agnostic to this specific partitioning scheme. The degradation engine can be readily adapted to accommodate alternative distortion sets or categorizations as dictated by the specific requirements of different application domains.

\subsection{Pretraining Optimization and Hyperparameters}
The perceptual branch is pretrained for 100 epochs with a global batch size of 384 using the MoCov3 \cite{MoCov3} framework. We employ the AdamW \cite{AdamW} optimizer with a peak learning rate of $1 \times 10^{-4}$ and starting weight decay of $0.001$. The teacher networks are updated using an exponential moving average of the weights of the student encoder. We employ a warmup cosine learning rate schedule with a linear warmup for the first 1000 steps, starting at a learning rate of $1 \times 10^{-6}$ and ultimately decaying to a final minimum learning rate of $1 \times 10^{-6}$. For weight decay and momentum, we also utilize a cosine schedule that gradually anneals the weight decay coefficient from $0.001$ to $0.1$ and the momentum coefficient from $0.994$ to $1.0$ over the duration of training. The contrastive temperature $\tau$ is set to $0.1$. The geometric threshold $\gamma$ for resolving structural conflicts is set to $0.05$. To ensure training stability and prevent gradient explosions, gradient clipping is applied with a maximum norm of $10.0$. Each degradation is encoded to a degradation label, while mask ratios (normalized to [0, 1]) are concatenated to form a constant-length vector ($-1$ padding is applied to maintain a fixed length). We employ the square-root-256 rule to scale the learning rate based on the batch size and the number of GPUs used for training. We employ horizontal flip and random cropping as our only data augmentations before passing the images through the degradation engine. We employ NVIDIA A100 and RTX 4090 GPUs for our experiments. The training time for each pretraining setting is around 6 to 8 hours. Following pretraining, SLIDE-IQA concatenates features from the original resolution and a downsampled version (half) to form the final feature representation, following prior works such as CONTRIQUE \cite{CONTRIQUE}, ReIQA \cite{ReIQA}, and TRIQA \cite{TRIQA}.

\subsection{Evaluation Protocols for NR-IQA Benchmarks}
We evaluate the NR-IQA performance of SLIDE-IQA on a comprehensive suite of eight standard benchmarks, including authentic and synthetic distortion datasets. The following protocols are followed for each dataset:
\begin{itemize}
	\item \textbf{KonIQ} \cite{KonIQ}: The dataset contains 10073 images with authentic distortions. We follow the standard protocol of using 80\% of the data for training and 20\% for testing.
	\item \textbf{CLIVE} \cite{CLIVE}: The dataset contains around 1162 images with authentic distortions. We follow the standard protocol of using 80\% of the data for training and 20\% for testing.
	\item \textbf{FLIVE} \cite{FLIVE}: The dataset contains nearly 40000 real-world images with authentic distortions, with each having variable spatial dimensions. We follow the standard protocol mentioned in the paper \cite{FLIVE}, i.e., images with at least one dimension more than 640 pixels are considered in the test dataset, and the rest are used for training. Images are padded and cropped to a resolution of 640x640 for both training and evaluation.
	\item \textbf{SPAQ} \cite{SPAQ}: The dataset contains 11125 images with authentic distortions. The resolutions of images in the dataset vary widely. Hence, we follow the standard protocol mentioned in the paper \cite{SPAQ}, i.e., we resize the images before evaluation such that the shorter side is 512.
	\item \textbf{LIVE-IQA} \cite{LIVE-IQA}: The dataset contains 29 reference images and 779 distorted images with synthetic distortions. We follow the standard protocol of using 80\% of the data for training and 20\% for testing.
	\item \textbf{CSIQ-IQA} \cite{CSIQ-IQA}: The dataset contains 30 reference images and 866 distorted images with synthetic distortions. We follow the standard protocol of using 80\% of the data for training and 20\% for testing.
	\item \textbf{TID-2013} \cite{TID-2013}: The dataset contains 25 reference images and 3000 distorted images with synthetic distortions. We follow the standard protocol of using 80\% of the data for training and 20\% for testing.
	\item \textbf{KADID-10k} \cite{KADID}: The dataset contains 81 reference images and 10125 distorted images with synthetic distortions. We follow the standard protocol of using 80\% of the data for training and 20\% for testing.
\end{itemize}
During training and evaluation on synthetic distortion datasets, we ensure that there is no source content overlap between the training and testing splits, i.e., the reference images used to generate the distorted images in the training set are completely disjoint from those used in the testing set. This ensures that the model is not simply memorizing content-specific cues but is truly learning to assess image quality based on the distortions present. Following the standard evaluation protocol for NR-IQA, we perform parameter search and evaluate the performance using Spearman's Rank Correlation Coefficient (SRCC) and Pearson's Linear Correlation Coefficient (PLCC) between the predicted quality scores and the ground truth Mean Opinion Scores (MOS). We repeat this process across 10 random train-test splits and report the median SRCC and PLCC to ensure a robust and reliable evaluation of the model's performance.

\subsection{Zero-Shot Performance}
\begin{figure}
	\centering
	\begin{subfigure}{0.32\textwidth}
		\centering
		\includegraphics[width=\textwidth]{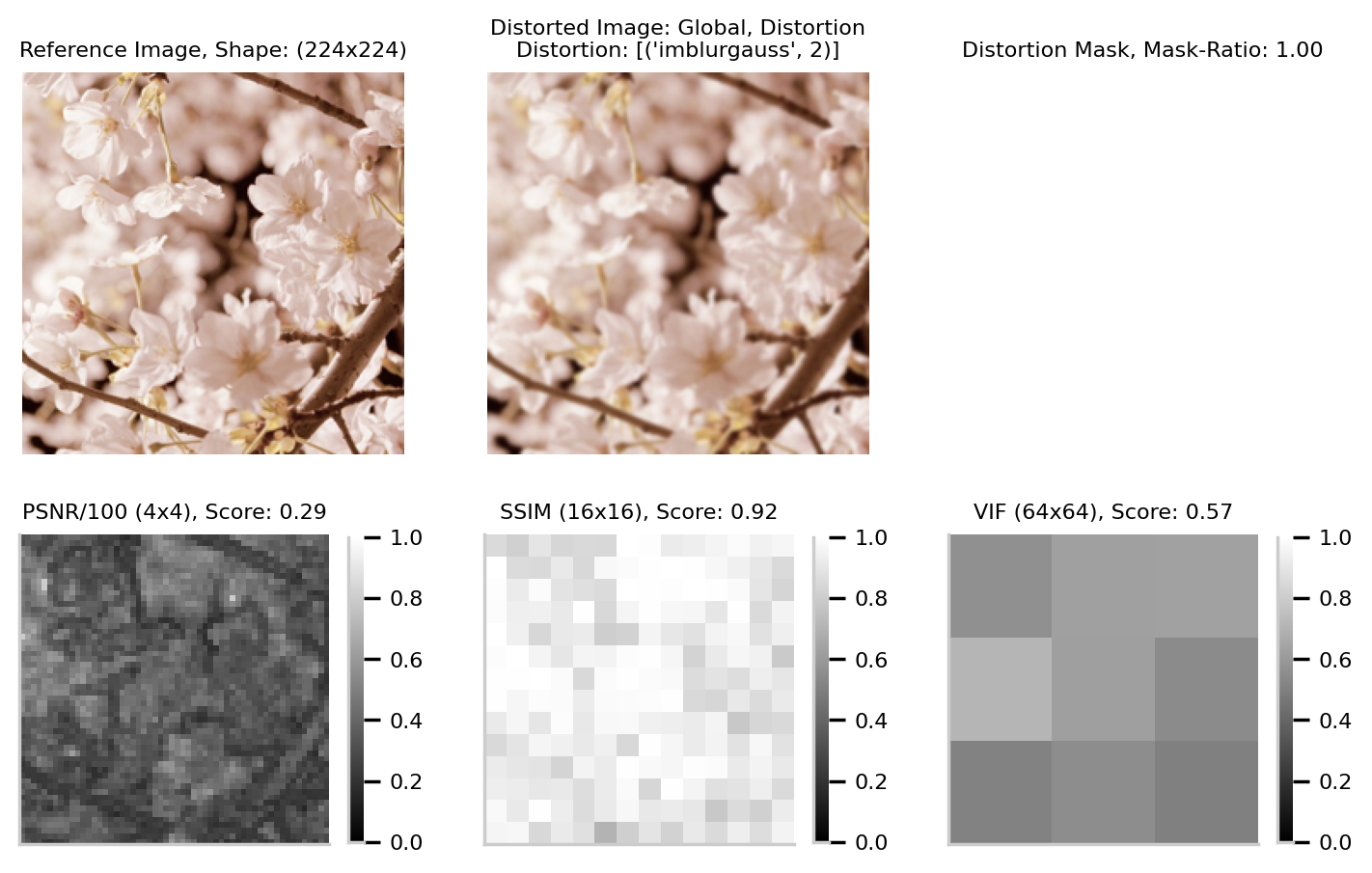}
	\end{subfigure}
	\hfill
	\begin{subfigure}{0.32\textwidth}
		\centering
		\includegraphics[width=\textwidth]{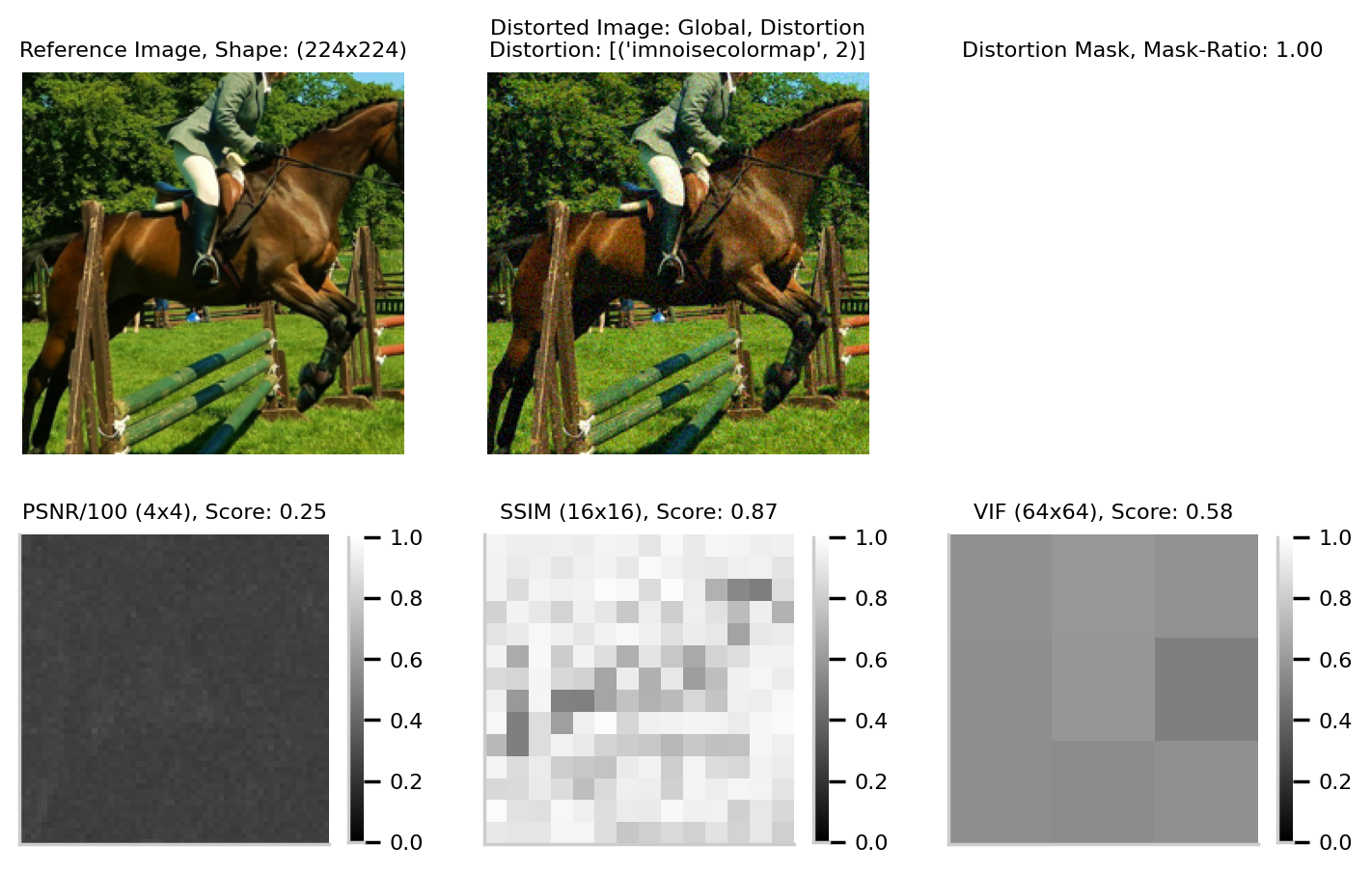}
	\end{subfigure}
	\hfill
	\begin{subfigure}{0.32\textwidth}
		\centering
		\includegraphics[width=\textwidth]{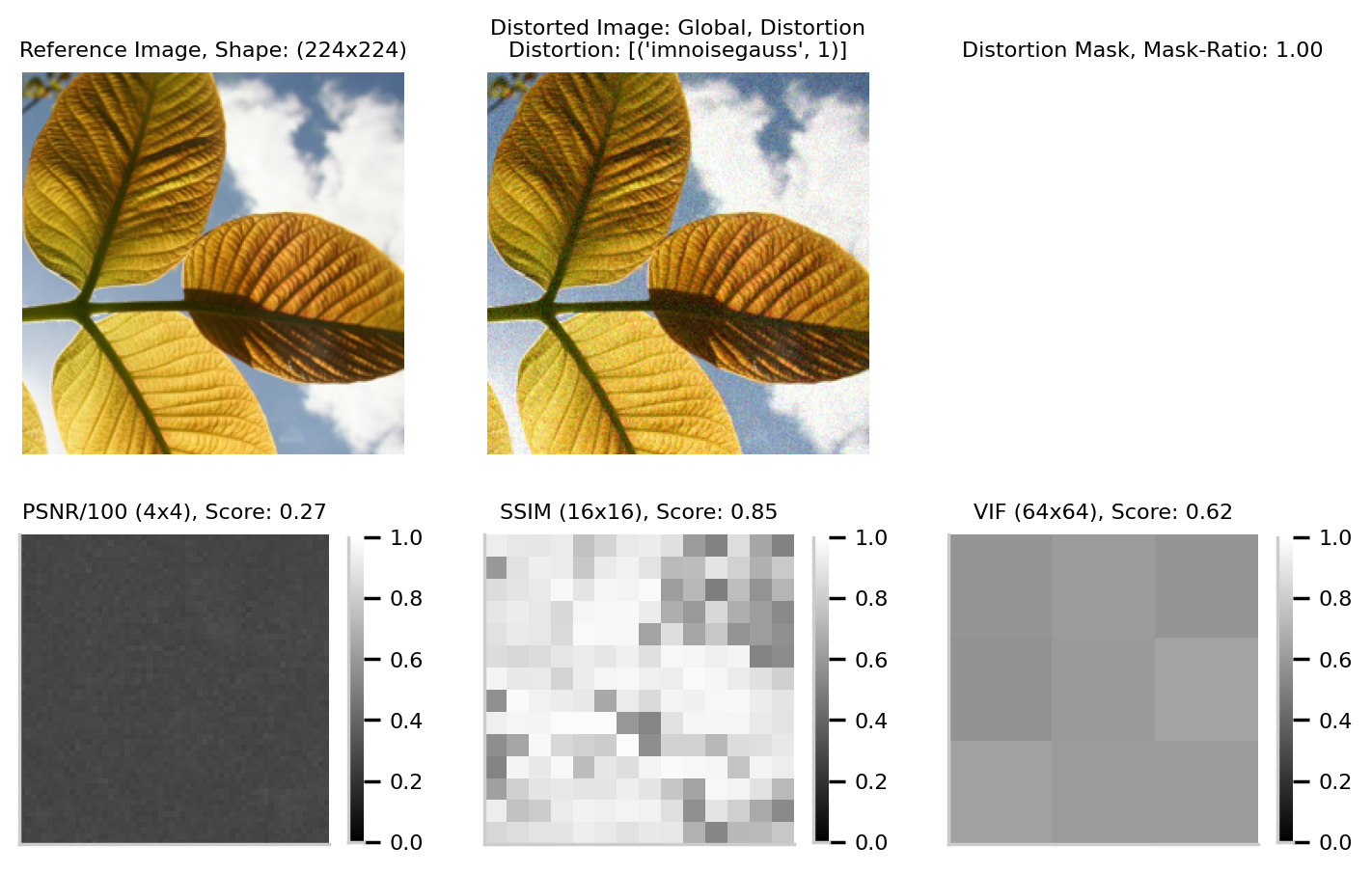}
	\end{subfigure}
	\begin{subfigure}{0.32\textwidth}
		\centering
		\includegraphics[width=\textwidth]{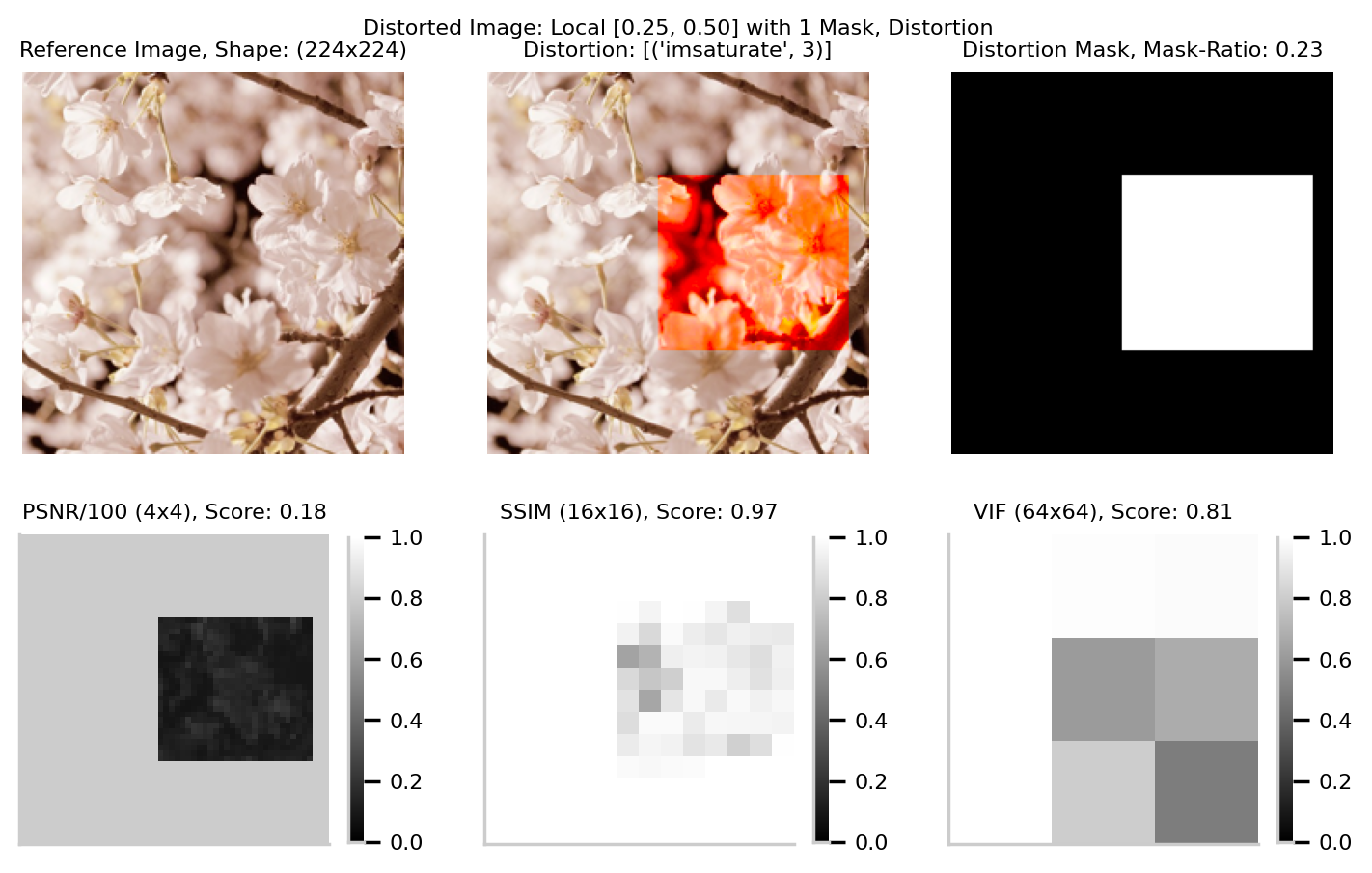}
	\end{subfigure}
	\hfill
	\begin{subfigure}{0.32\textwidth}
		\centering
		\includegraphics[width=\textwidth]{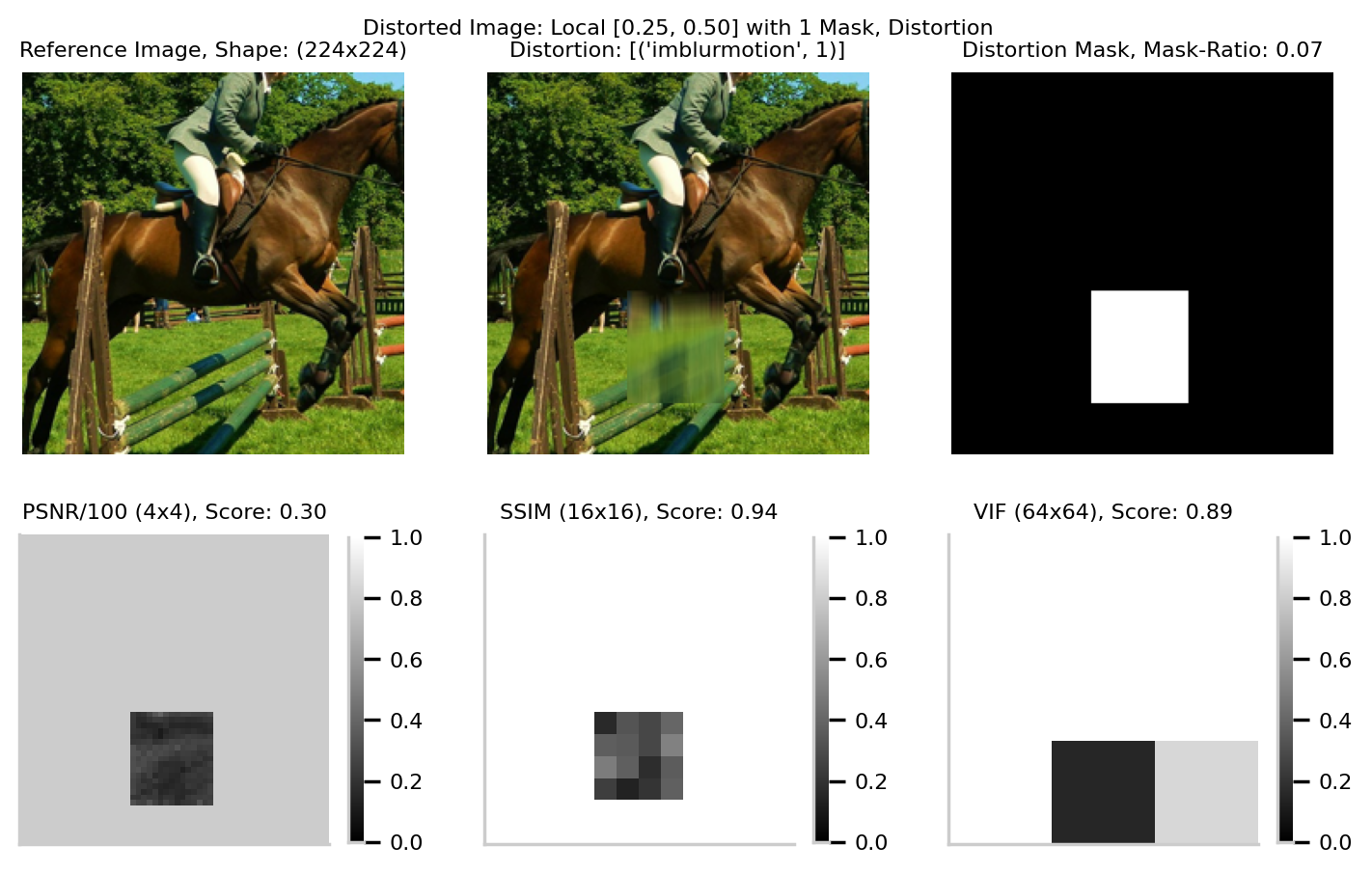}
	\end{subfigure}
	\hfill
	\begin{subfigure}{0.32\textwidth}
		\centering
		\includegraphics[width=\textwidth]{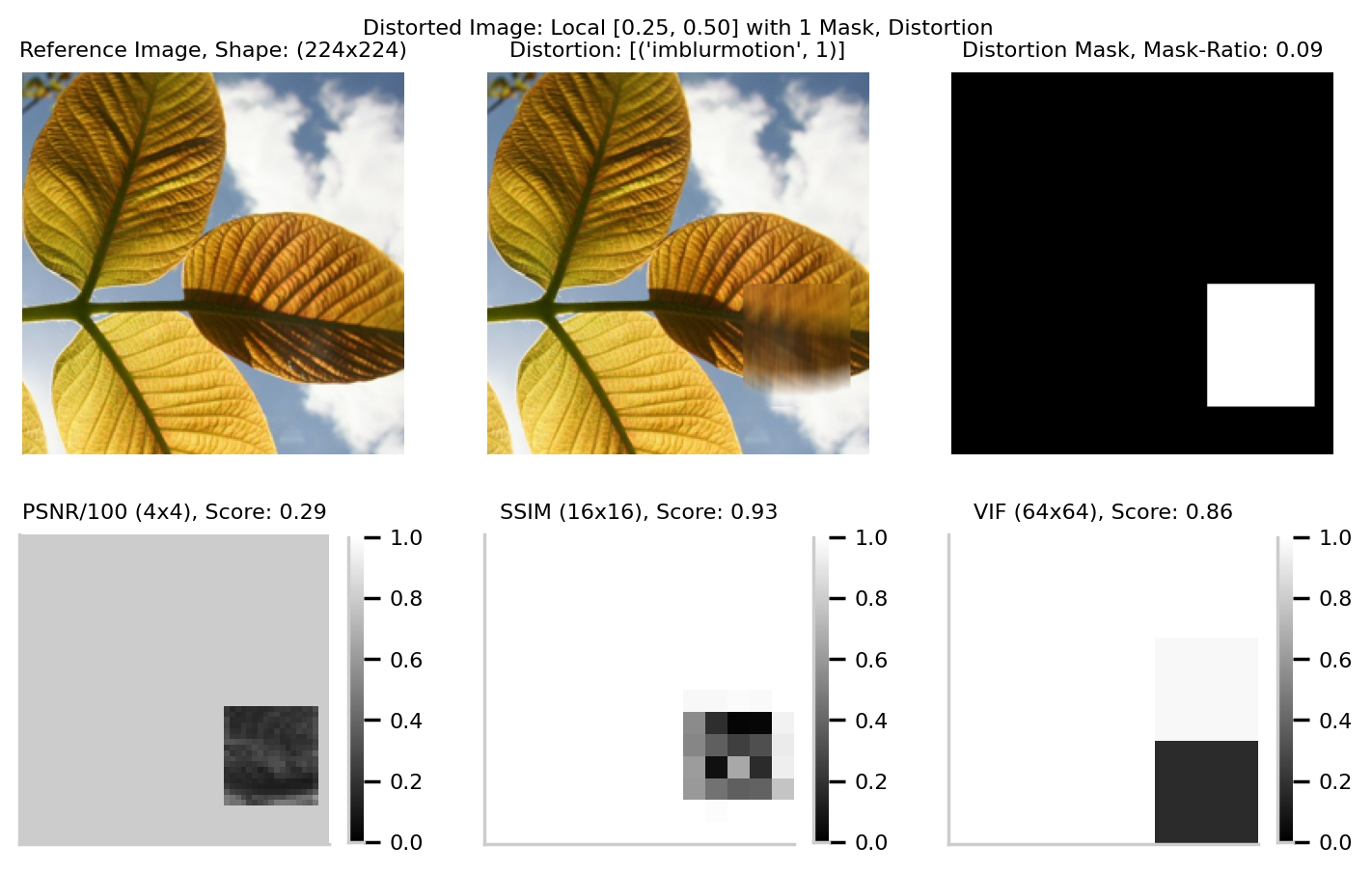}
	\end{subfigure}
	\begin{subfigure}{0.32\textwidth}
		\centering
		\includegraphics[width=\textwidth]{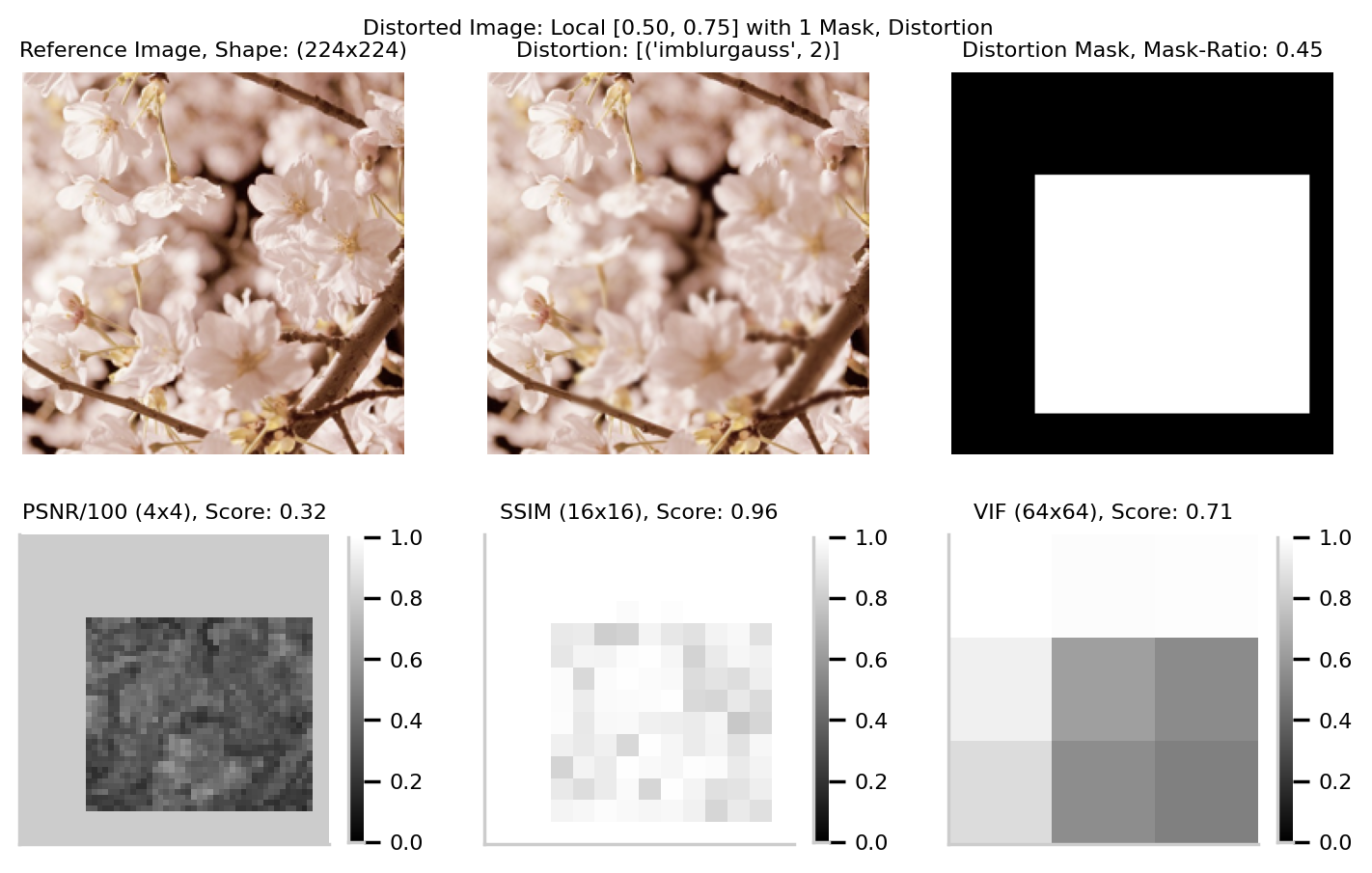}
	\end{subfigure}
	\hfill
	\begin{subfigure}{0.32\textwidth}
		\centering
		\includegraphics[width=\textwidth]{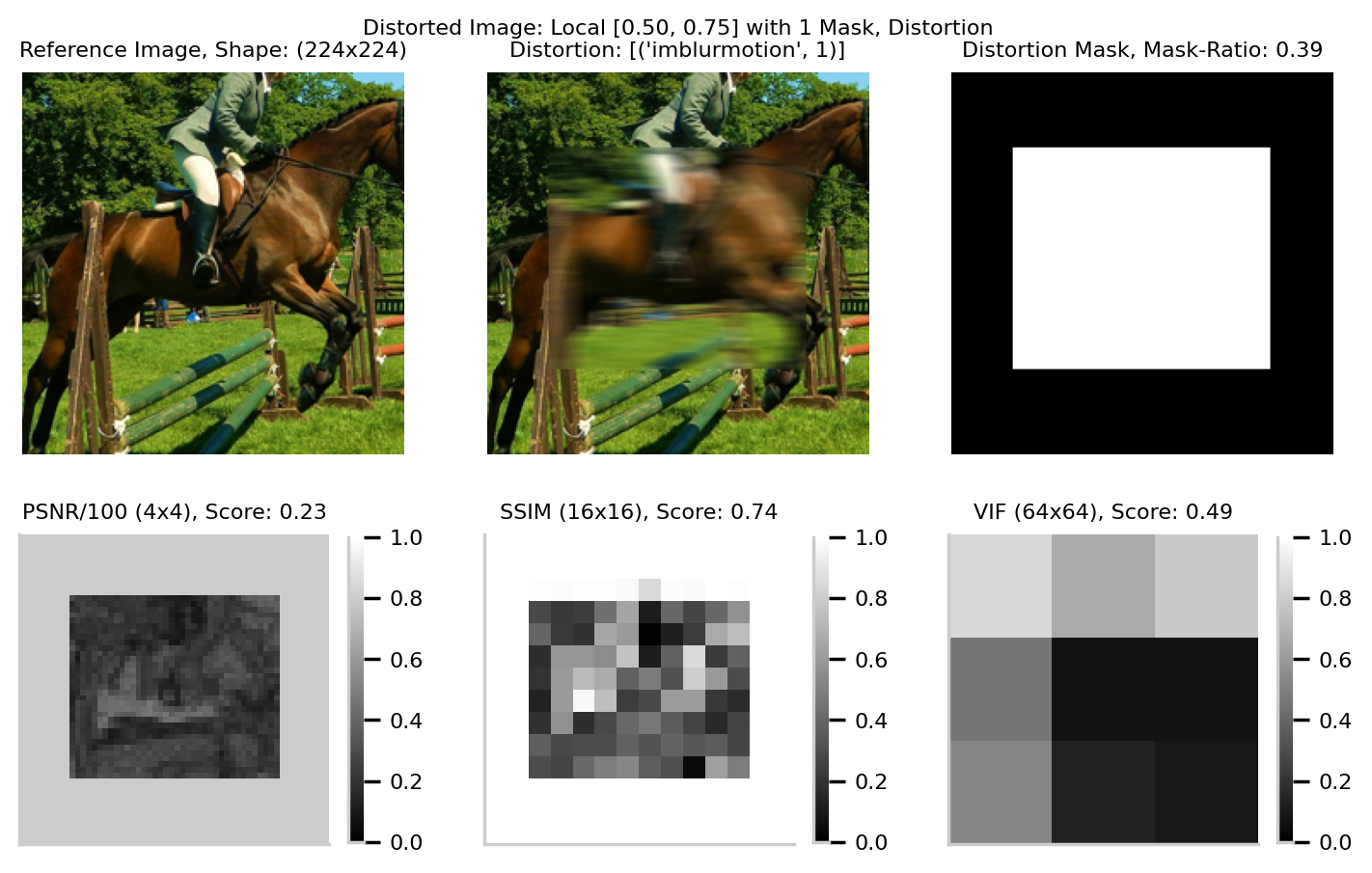}
	\end{subfigure}
	\hfill
	\begin{subfigure}{0.32\textwidth}
		\centering
		\includegraphics[width=\textwidth]{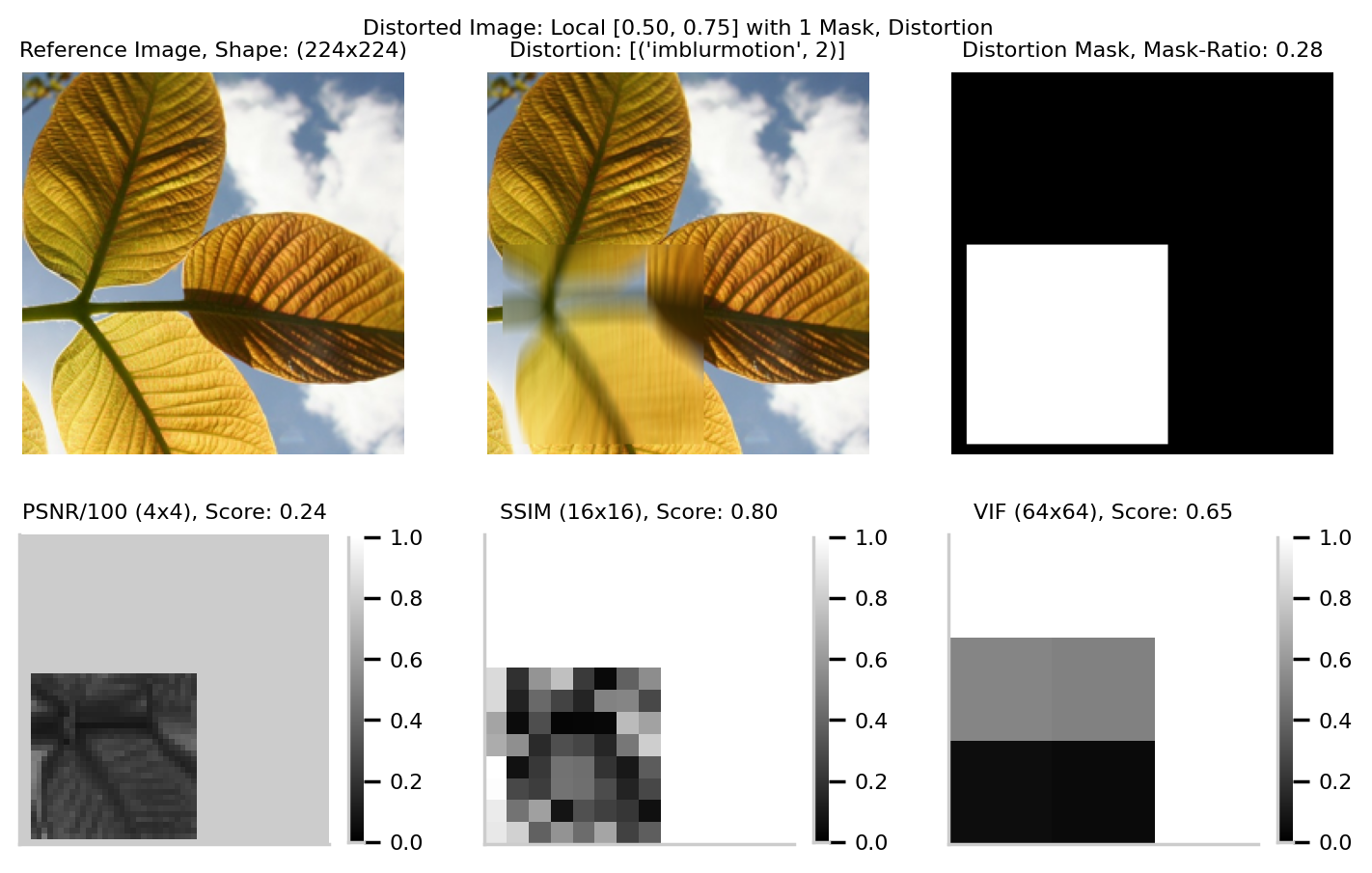}
	\end{subfigure}
	\begin{subfigure}{0.32\textwidth}
		\centering
		\includegraphics[width=\textwidth]{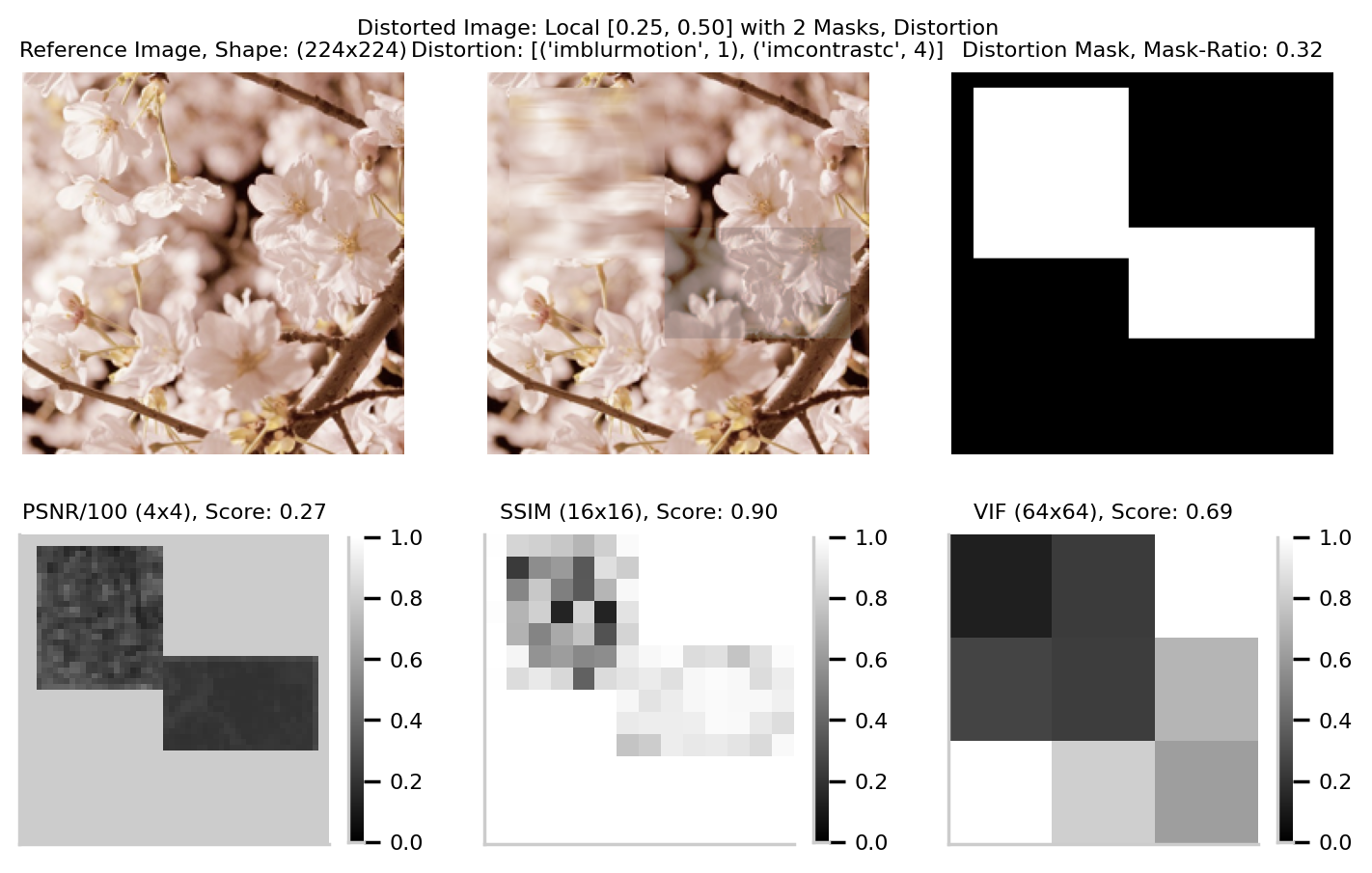}
	\end{subfigure}
	\hfill
	\begin{subfigure}{0.32\textwidth}
		\centering
		\includegraphics[width=\textwidth]{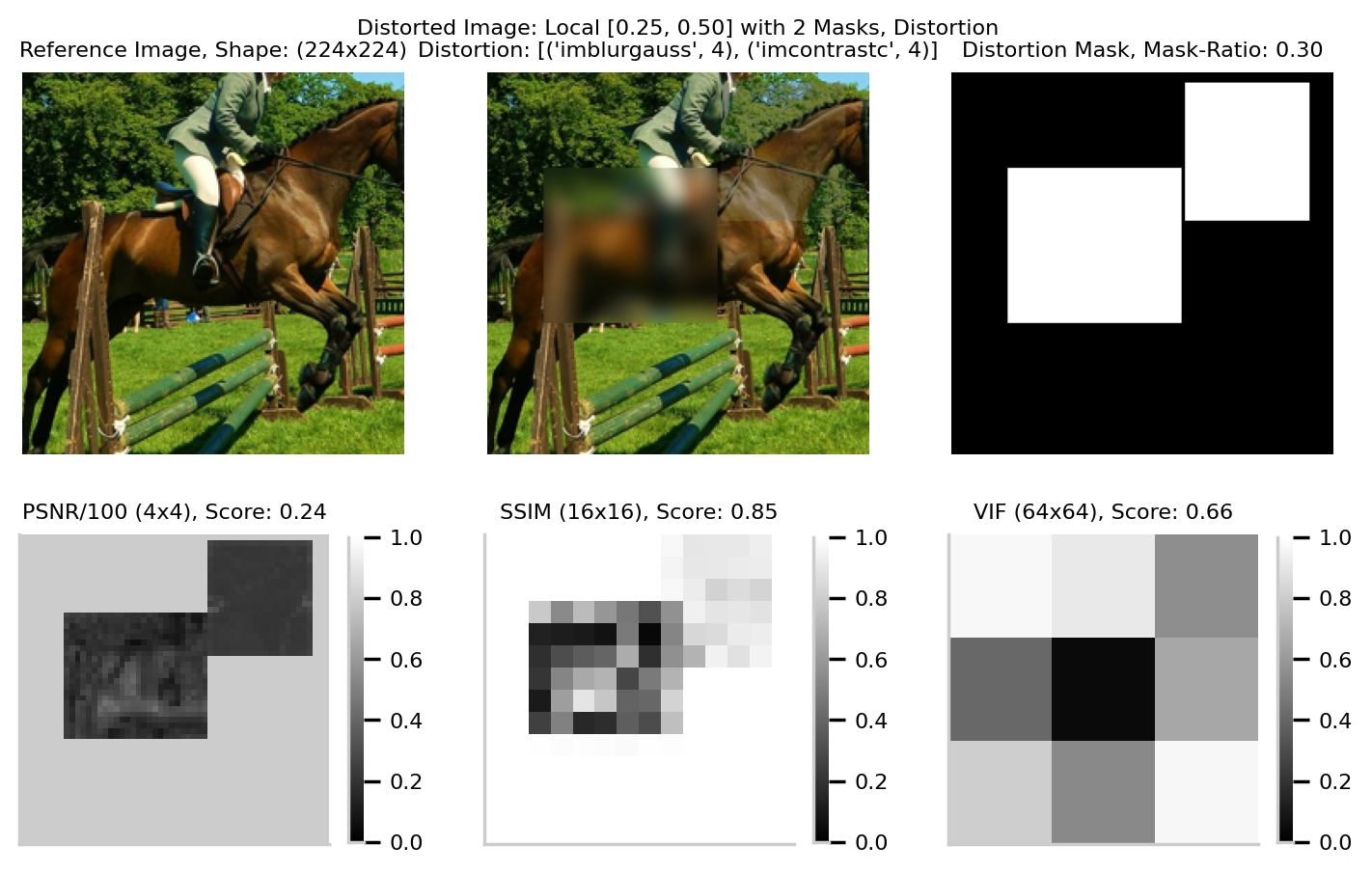}
	\end{subfigure}
	\hfill
	\begin{subfigure}{0.32\textwidth}
		\centering
		\includegraphics[width=\textwidth]{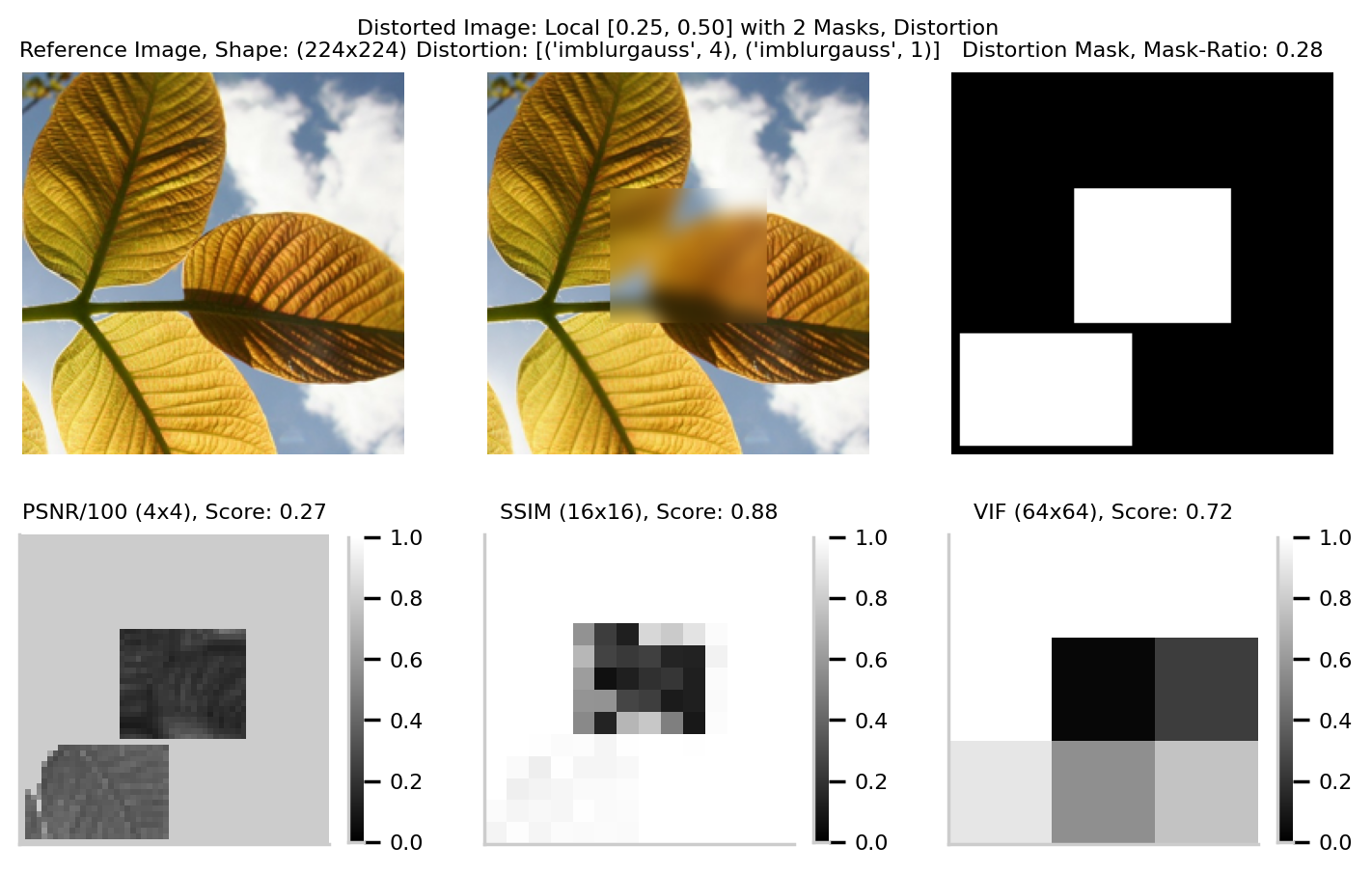}
	\end{subfigure}
	\begin{subfigure}{0.32\textwidth}
		\centering
		\includegraphics[width=\textwidth]{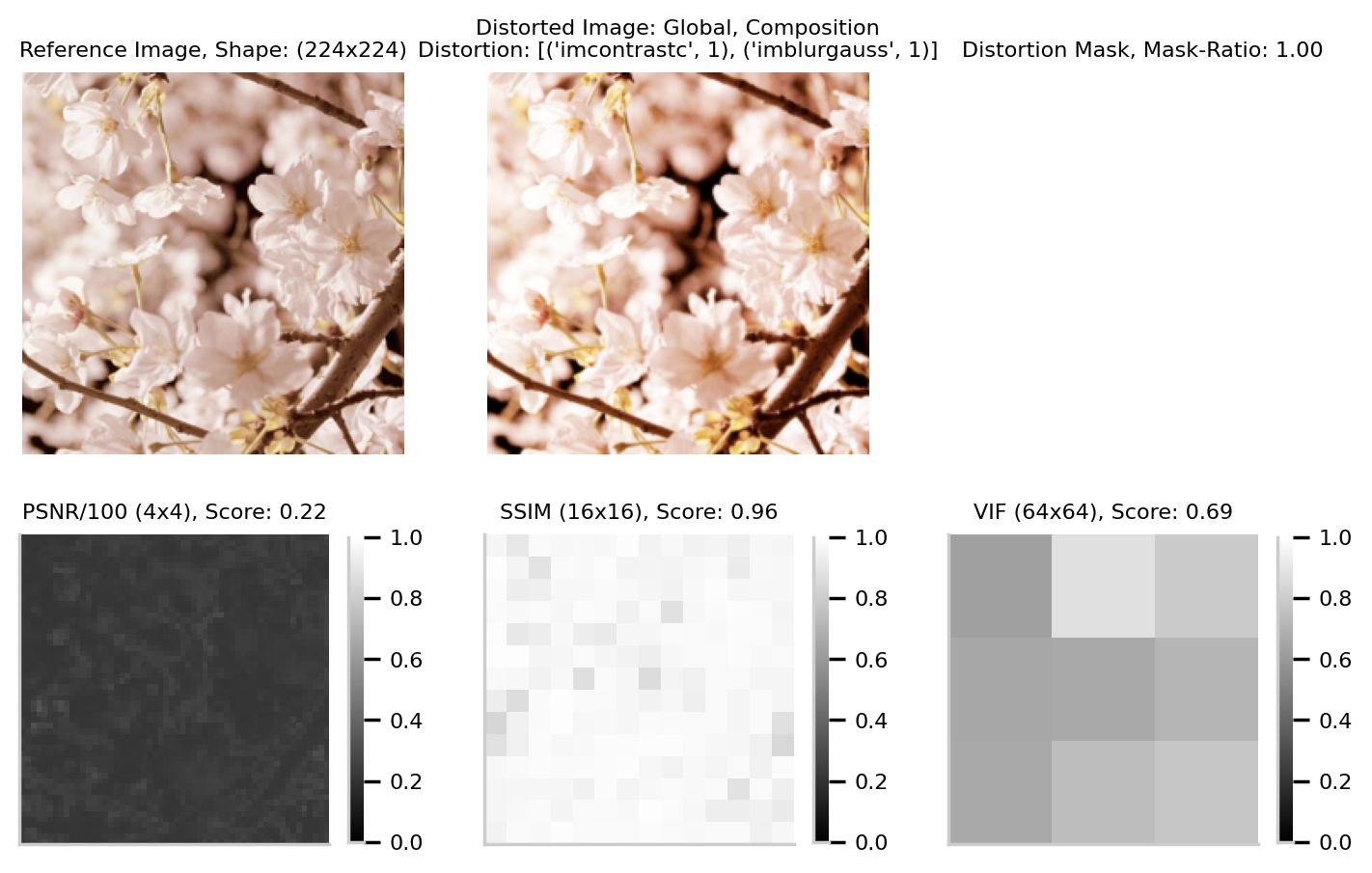}
	\end{subfigure}
	\hfill
	\begin{subfigure}{0.32\textwidth}
		\centering
		\includegraphics[width=\textwidth]{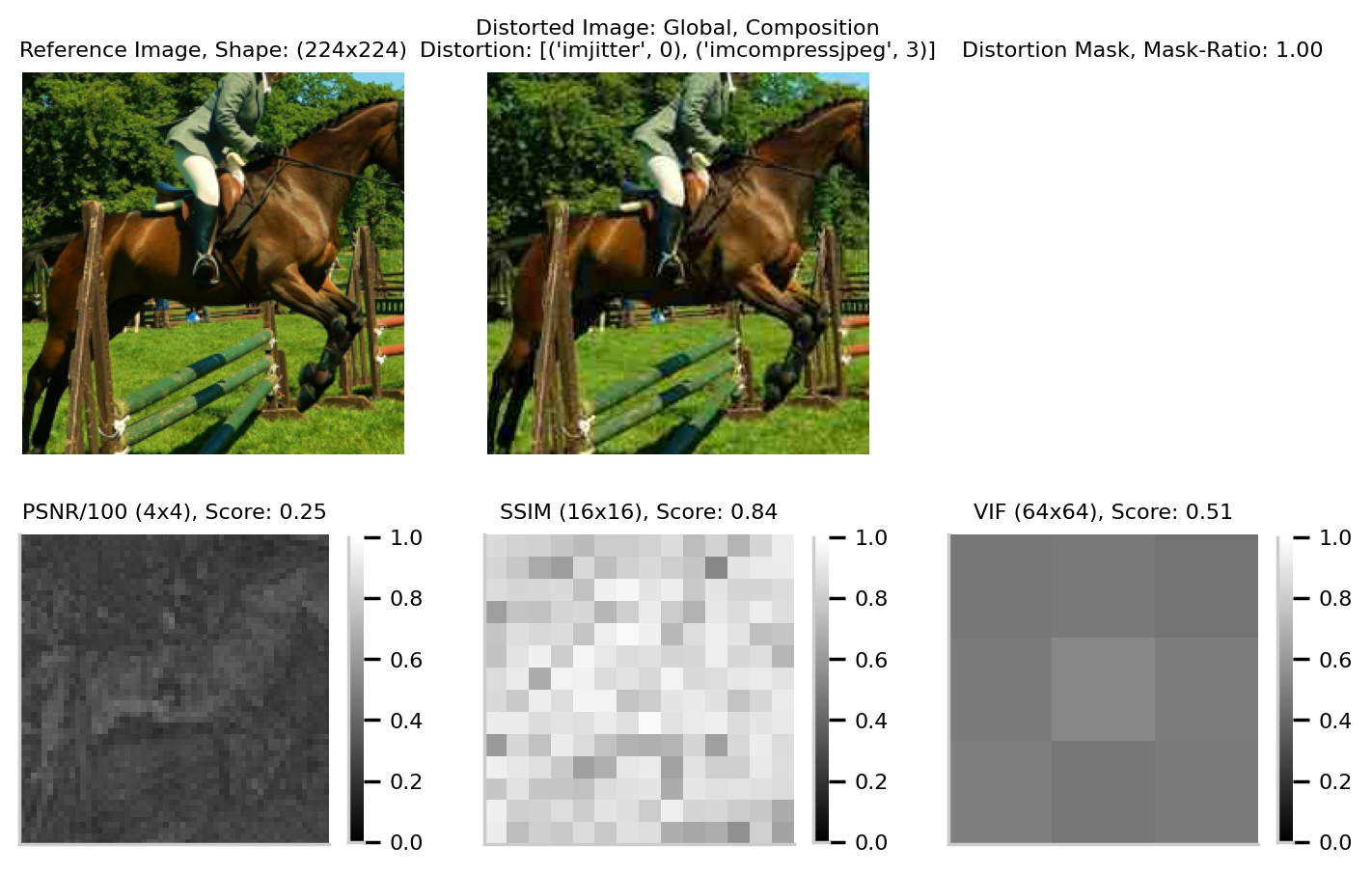}
	\end{subfigure}
	\hfill
	\begin{subfigure}{0.32\textwidth}
		\centering
		\includegraphics[width=\textwidth]{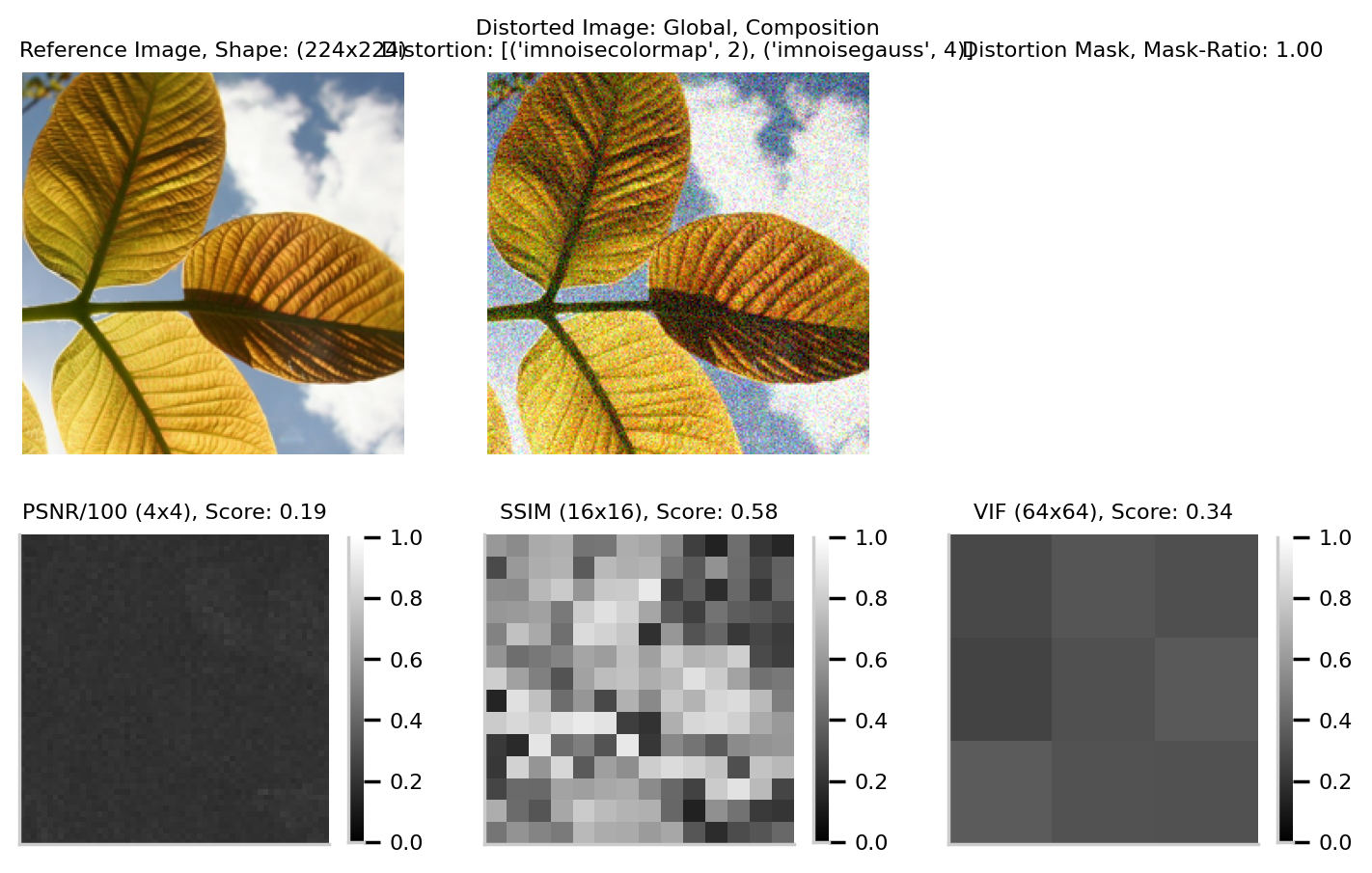}
	\end{subfigure}
	\caption{Visualization of quality score maps at the patch level from various FR-IQA models on the diagnostic test datasets.}
	\label{fig:quality_score_maps}
\end{figure}

To isolate the architectural origin of this insensitivity, we extract patch-level spatial quality maps from PSNR \cite{PSNR}, SSIM \cite{SSIM}, and VIF \cite{VIF}. As illustrated in Figure \ref{fig:quality_score_maps}, we compute localized metrics across non-overlapping spatial windows: 4x4 (PSNR), 16x16 (SSIM), and 64x64 (VIF). These quality score maps are normalized to [0, 1] to facilitate cross-metric comparison. The quality score computed using the entire image is shown in the title of each heatmap. It can be observed that at the individual patch level, metrics such as SSIM and VIF successfully detect localized distortions, assigning steep penalties to the perceptually degraded regions. However, this regional fidelity is lost in the quality scores computed across larger spatial windows, i.e., the entire image, resulting in scores towards the pristine end of the spectrum. 

\subsection{Ablation Studies}
\begin{figure}[!ht]
	\centering
	\begin{subfigure}{0.24\textwidth}
		\centering
		\includegraphics[width=\textwidth]{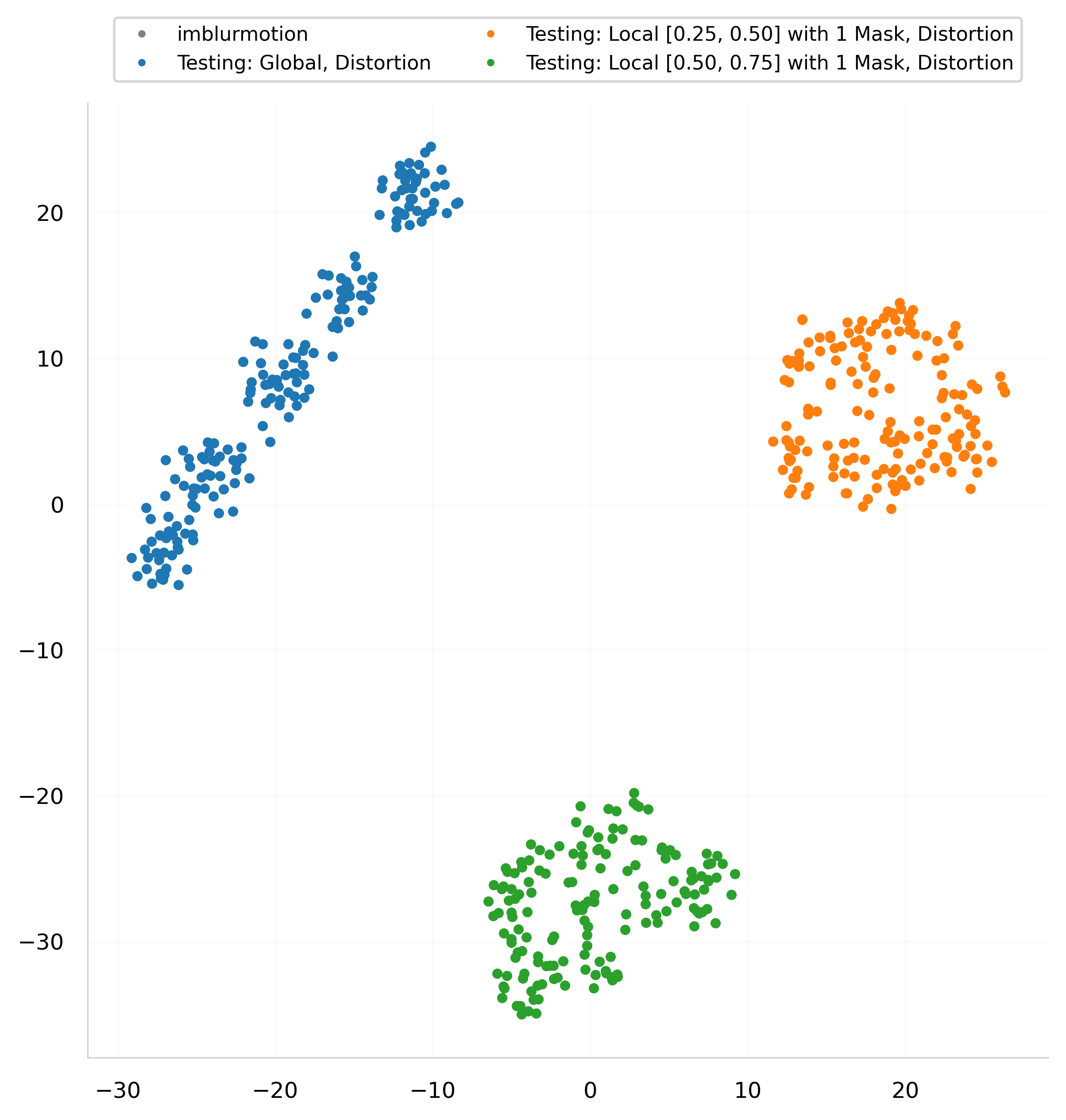}
		\caption{$\gamma = 0.0$}
	\end{subfigure}
	\hfill
	\begin{subfigure}{0.24\textwidth}
		\centering
		\includegraphics[width=\textwidth]{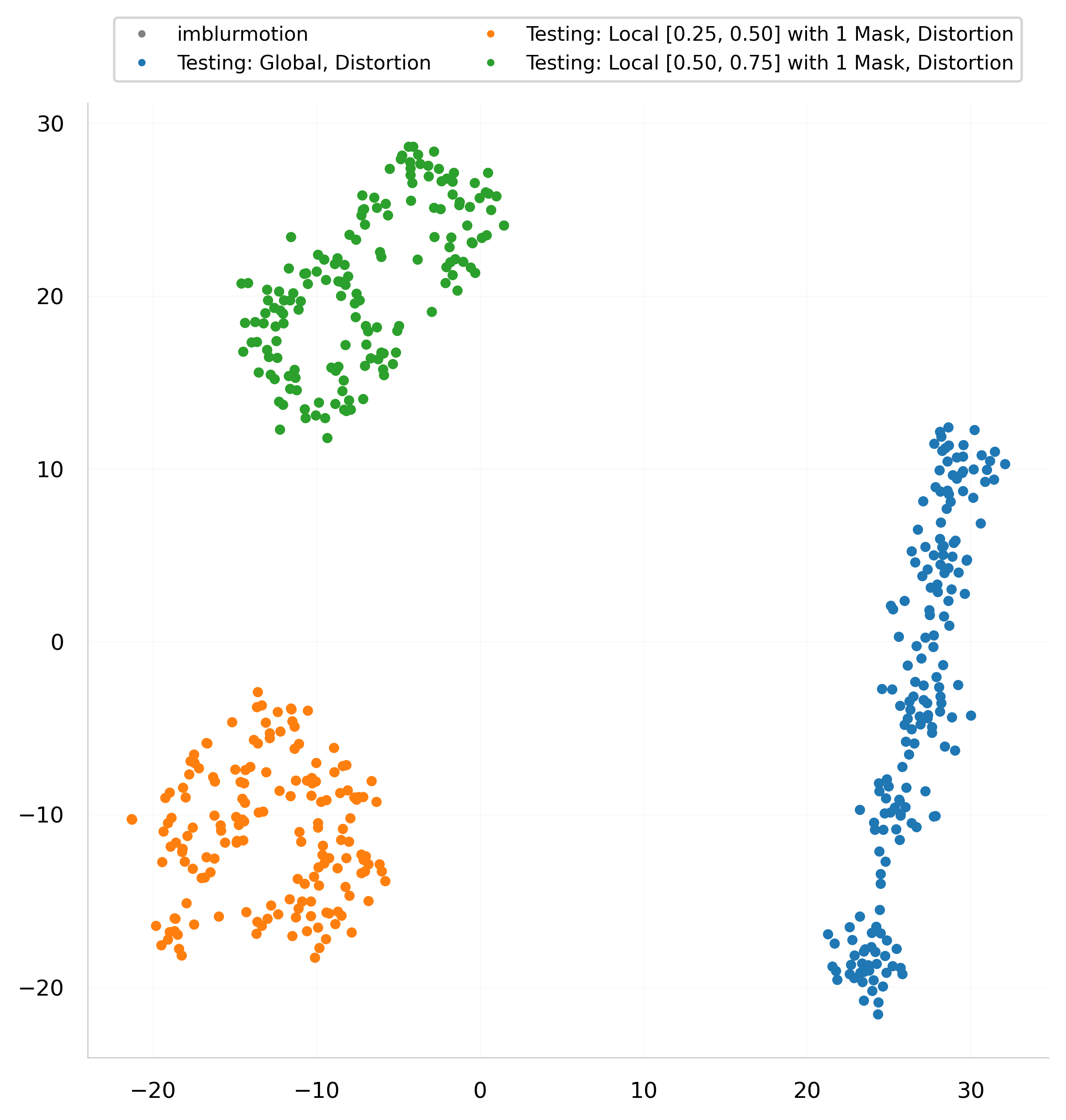}
		\caption{$\gamma = 0.05$}
	\end{subfigure}
	\hfill
	\begin{subfigure}{0.24\textwidth}
		\centering
		\includegraphics[width=\textwidth]{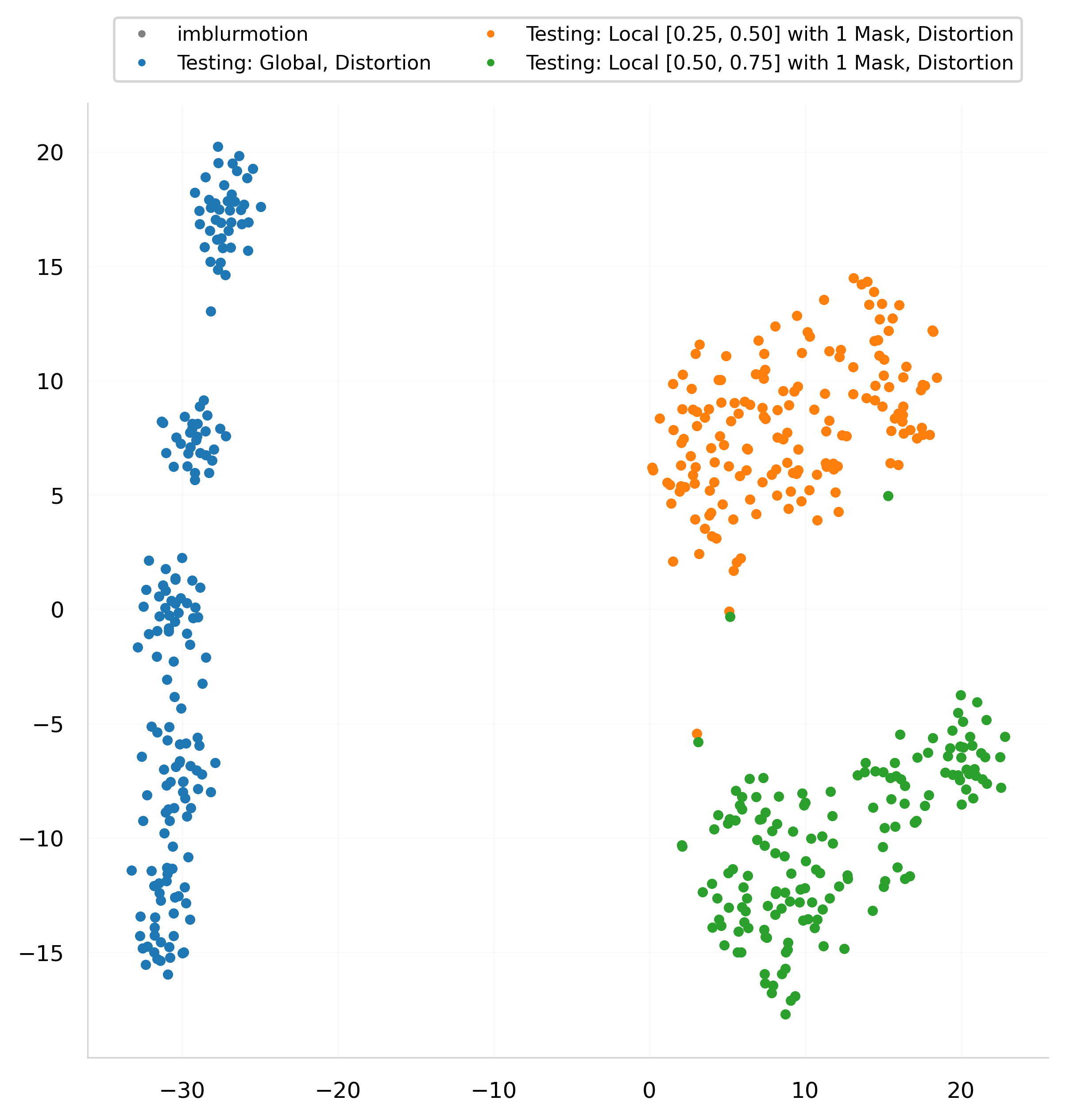}
		\caption{$\gamma = 0.15$}
	\end{subfigure}
	\hfill
	\begin{subfigure}{0.24\textwidth}
		\centering
		\includegraphics[width=\textwidth]{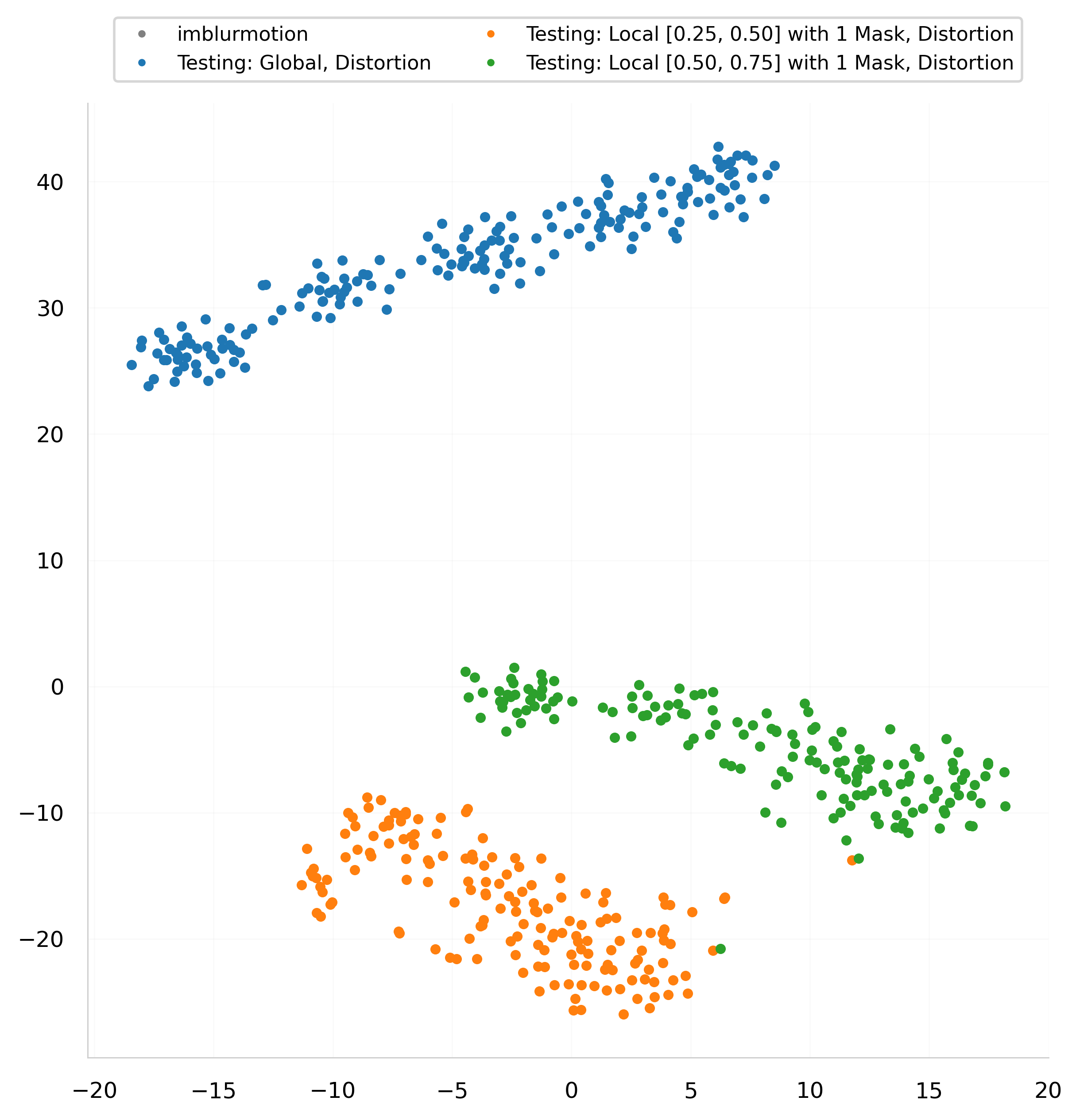}
		\caption{$\gamma = \infty$}
	\end{subfigure}
	\caption{t-SNE visualization of the representations from the perceptual branch pretrained with different values of the geometric threshold $\gamma$.}
	\label{fig:gamma_ablation}
\end{figure}

Figure \ref{fig:gamma_ablation} shows the t-SNE visualization of the representations from the perceptual branch pretrained with different values of the geometric threshold $\gamma$ for motion blur distortion. As $\gamma$ increases, the representations of the distorted images exhibit greater dispersion. This occurs because higher values of $\gamma$ reduce the number of samples with identical degradations assigned to the ignored set, instead forcing them to be treated as positive pairs. This compels the model to discard the spatial scale (i.e., mask ratios) and optimize exclusively for the physical distortion type. Consequently, this results in a uniform distribution of features across varying mask sizes. On the other hand, lower values of $\gamma$ aggressively exclude spatially divergent samples with similar degradation during the loss computation. By silencing these gradient collisions, the model is permitted to pull samples with smaller mask ratio differences closer together, while ignoring the gradient signals from samples with larger mask ratio differences, resulting in a more tightly clustered representation space. Given the large batch size and the relatively comparable number of image degradations, we observe that changes to $\gamma$ have a relatively minimal effect on the overall performance of the model on NR-IQA benchmarks as shown in Table \ref{tab:gamma-ablation}, as each anchor representation of the student encoder sees around 3.66 positives, 710 negatives, and 0.56 ignored teacher encoder representations on average per batch for $\gamma = 0.05$.

\begin{table*}[!ht]
	\centering
	\scriptsize
	\setlength{\tabcolsep}{4pt}
	\renewcommand{\arraystretch}{1.02}
	\resizebox{\textwidth}{!}{%
	\begin{tabular}{@{}l *{14}{c}@{}}
		\midrule
		\multirow{3}{*}{SLIDE-IQA-Q} & \multicolumn{6}{c}{Authentic Distortions --- ``Images in the Wild''} & \multicolumn{8}{c}{Synthetic Distortions} \\
		\cmidrule(lr){2-7} \cmidrule(lr){8-15}
		& \multicolumn{2}{c}{KonIQ} & \multicolumn{2}{c}{CLIVE} & \multicolumn{2}{c}{SPAQ} & \multicolumn{2}{c}{LIVE-IQA} & \multicolumn{2}{c}{CSIQ-IQA} & \multicolumn{2}{c}{TID-2013} & \multicolumn{2}{c}{KADID} \\
		\cmidrule(lr){2-3} \cmidrule(lr){4-5} \cmidrule(lr){6-7} \cmidrule(lr){8-9} \cmidrule(lr){10-11} \cmidrule(lr){12-13} \cmidrule(lr){14-15}
		& SRCC & PLCC & SRCC & PLCC & SRCC & PLCC & SRCC & PLCC & SRCC & PLCC & SRCC & PLCC & SRCC & PLCC \\
		\midrule
		$\gamma = 0.0$ & 0.840 & 0.858 & \textbf{0.727} & \textbf{0.733} & 0.879 & 0.885 & 0.955 & 0.958 & 0.945 & 0.958 & 0.885 & 0.89 & 0.928 & 0.932 \\
		$\gamma = 0.05$ & \textbf{0.854} & \textbf{0.869} & 0.690 & 0.710 & 0.885 & 0.89 & 0.955 & \textbf{0.961} & 0.940 & 0.954 & \textbf{0.881} & \textbf{0.892} & \textbf{0.932} & \textbf{0.934} \\
		$\gamma = 0.15$ & 0.851 & 0.866 & 0.689 & 0.724 & 0.886 & 0.891 & 0.955 & 0.959 & 0.941 & 0.955 & 0.880 & 0.891 & 0.927 & 0.930 \\
		$\gamma = \infty$ & 0.850 & 0.864 & 0.696 & 0.712 & 0.88 & 0.886 & \textbf{0.957} & 0.960 & \textbf{0.946} & \textbf{0.960} & 0.880 & 0.89 & 0.929 & 0.932 \\
		\bottomrule
		\end{tabular}
 }
	\caption{Performance of SLIDE-IQA-Q for different values of $\gamma$ on authentic and synthetic distortion datasets.}
	\label{tab:gamma-ablation}
\end{table*}

\begin{figure}[!ht]
	\centering
	\begin{subfigure}{0.95\textwidth}
		\centering
		\includegraphics[width=\textwidth]{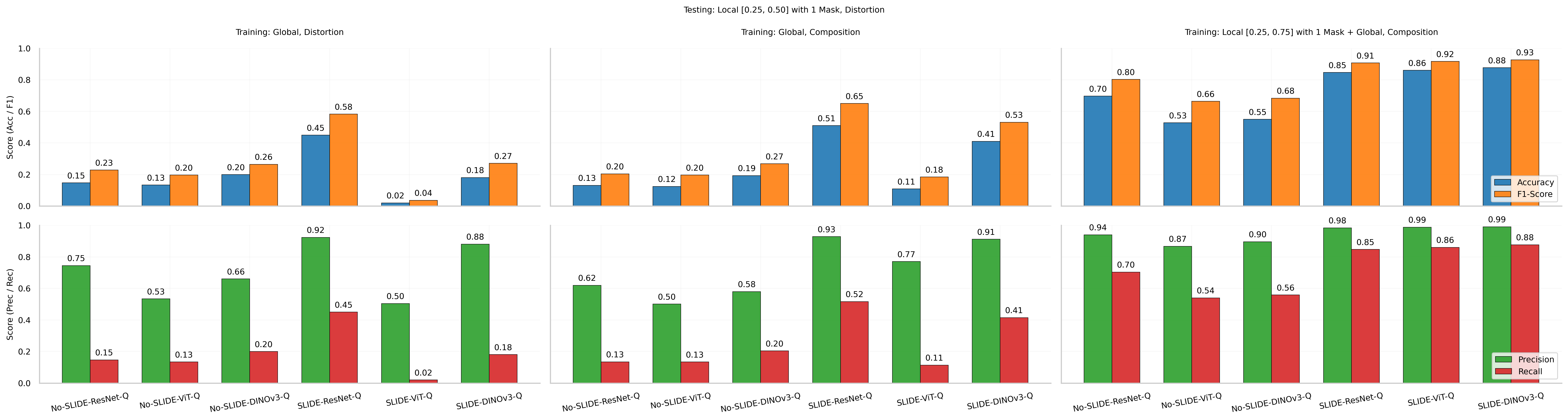}
		\caption{Local[0.25, 0.50] + 1 Mask}
	\end{subfigure}
	\begin{subfigure}{0.95\textwidth}
		\centering
		\includegraphics[width=\textwidth]{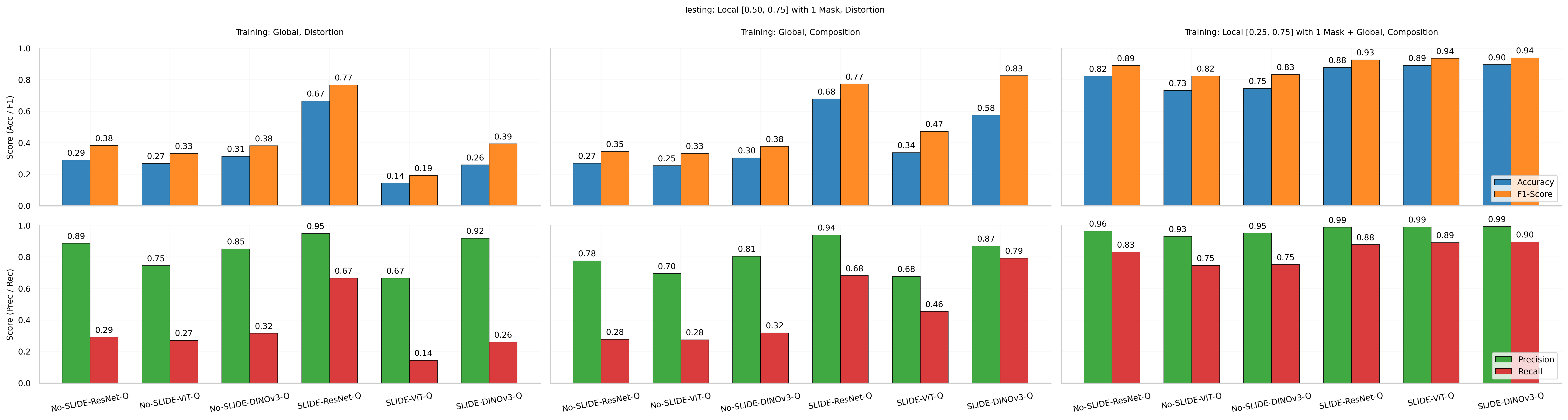}
		\caption{Local[0.50, 0.75] + 1 Mask}
	\end{subfigure}
	\caption{Comparison of the performance of linear probes trained under different training regimes for various model architectures on the diagnostic test datasets.}
	\label{fig:training_regime_comparison}
\end{figure}

We further study the performance of linear probes trained under different training regimes for various model architectures on the diagnostic test datasets. To demonstrate the efficacy of our proposed pretraining framework, we design a `No-SLIDE-IQA' baseline where, during pretraining, the provided masks are ignored for all the distortions, i.e., all the distortions are applied globally. To isolate the effect of image encoder architecture and initialization, we experiment with three different model architectures: ResNet50 \cite{ResNet50} pretrained on ImageNet-1K \cite{ImageNet} using supervised learning, ViT-S/16 \cite{ViT} pretrained on ImageNet-1K using supervised learning, and DINOv3 ViT-S/16 \cite{DINOv3}. The linear probes are trained under three different training regimes: `Global Distortion', `Global Composition', and `Global + Local[0.25, 0.75]'. The performance of the linear probes is evaluated on the diagnostic test datasets: `Local[0.25, 0.50] + 1 Mask' and `Local[0.50, 0.75] + 1 Mask', and the results are shown in Figure \ref{fig:training_regime_comparison}.

The `No-SLIDE-IQA' baseline performs significantly worse than models pretrained with the SLIDE-IQA framework consistently across most of the training regimes, model architectures, and diagnostic test datasets. This demonstrates the efficacy of our proposed pretraining framework in learning representations that are sensitive to localized distortions. Independent of the model architecture and initialization, image encoders pretrained with `No-SLIDE-IQA' perform similarly among themselves. These results suggest that the performance on the diagnostic test datasets is not simply a function of the model architecture or initialization, but the performance is significantly influenced by the training regime employed during pretraining of image encoders.

Among image encoders trained on the SLIDE-IQA framework, it can also be observed that the linear probes trained with the `Global Distortion' and `Global Composition' training regimes perform significantly worse than those trained with the `Global + Local[0.25, 0.75] + 1 Mask' training regime across all model architectures and diagnostic test datasets. ResNet50 \cite{ResNet50} trained on ImageNet-1K using supervised learning demonstrates better performance than ViTs on these training regimes. However, with the `Global + Local[0.25, 0.75] + 1 Mask' training regime, the performance of the linear probes trained on ViT-S/16 architecture significantly improves, surpassing that of ResNet50. Additionally, DINOv3 ViT-S/16 \cite{DINOv3} demonstrates better performance than the supervised ViT-S/16 \cite{ViT} across all training regimes and diagnostic test datasets. 

% References
{
    \small
    \bibliographystyle{plainnat}
    \bibliography{refs}
}

% Checklist
% \input{checklist}

\end{document}